\documentclass[sigconf,nonacm]{acmart}
\AtBeginDocument{
  }

\setcopyright{none}
\renewcommand\footnotetextcopyrightpermission[1]{}
\copyrightyear{2027}
\acmYear{2027}
\acmDOI{}
\acmConference[KDD '27]{Proceedings of the ACM SIGKDD Conference on Knowledge Discovery and Data Mining}{August 2027}{}
\acmISBN{}

\usepackage{tabularx}
\usepackage{multirow}
\usepackage{subcaption}
\usepackage{float}
\usepackage{algorithm}
\usepackage{algorithmic}
\usepackage{xcolor}
\usepackage{enumitem}
\begin{document}

\title{MM-ISTS: Cooperating Irregularly Sampled Time Series Forecasting with Multimodal Vision-Text LLMs}

\author{Zhi Lei}
\authornote{Both authors contributed equally to this paper.}
\email{zhilei@stu.ecnu.edu.cn}
\affiliation{%
  \institution{East China Normal University}
  \city{Shanghai}
  \country{China}
}

\author{Chenxi Liu}
\authornotemark[1]
\email{chenxi.liu@cair-cas.org.hk}
\affiliation{%
  \institution{Centre for Artificial Intelligence and Robotics, Hong Kong Institute of Science \& Innovation, Chinese Academy of Sciences}
  \city{Hong Kong}
  \country{China}
}

\author{Hao Miao}
\authornote{Corresponding authors.}
\email{hao.miao@polyu.edu.hk}
\affiliation{%
  \institution{Department of Computing, The Hong Kong Polytechnic University}
  \city{Hong Kong}
  \country{China}
}

\author{Wanghui Qiu}
\email{onehui@stu.ecnu.edu.cn}
\affiliation{%
  \institution{East China Normal University}
  \city{Shanghai}
  \country{China}
}

\author{Bin Yang}
\email{byang@dase.ecnu.edu.cn}
\affiliation{%
  \institution{East China Normal University}
  \city{Shanghai}
  \country{China}
}

\author{Chenjuan Guo}
\authornotemark[2]
\email{cjguo@dase.ecnu.edu.cn}
\affiliation{%
  \institution{East China Normal University}
  \city{Shanghai}
  \country{China}
}

\begin{abstract}
Irregularly sampled time series (ISTS) are widespread in real-world scenarios, exhibiting asynchronous observations on uneven time intervals across diverse variables. Existing ISTS forecasting methods often solely utilize historical observations to predict future ones while falling short in learning contextual semantics and fine-grained temporal patterns. To address these problems, we propose MM-ISTS, a multimodal ISTS forecasting framework augmented by vision-text large language models, which bridges temporal, visual, and textual modalities. MM-ISTS encompasses a two-stage encoding mechanism. In particular, a Cross-Modal Vision-Text Encoding module is proposed to automatically generate informative visual images and textual data, enabling the capture of intricate temporal patterns and comprehensive contextual understanding, in collaboration with multimodal LLMs (MLLMs). In parallel, ISTS encoding extracts complementary yet enriched temporal features from historical ISTS observations, including multi-view embedding fusion and a Temporal-Variable Encoder. Further, we propose an Adaptive Query-Based Feature Extractor to compress MLLM token embeddings while preserving useful knowledge, which in turn reduces computational costs. In addition, a Multimodal Alignment module with Modality-Aware Gating is designed to alleviate the modality gaps. Extensive experiments on real data offer insight into the effectiveness of the proposed solutions.
\end{abstract}

\ccsdesc[500]{Information systems~Spatial-temporal systems}

\keywords{ISTS Forecasting, Multimodal LLMs, Cross-Modal Alignment}

\maketitle

\section{Introduction}
The expanding instrumentation of processes throughout society with sensors yields a proliferation of time series data in various domains such as healthcare~\citep{goldberger2000physiobank,DBLP:conf/cinc/ReynaJSJSWSNC19}, transportation~\citep{DBLP:conf/aaai/Fan22,DBLP:conf/aaai/TangYSAMW20}, climate science~\citep{DBLP:conf/nips/RubanovaCD19,DBLP:conf/nips/BrouwerSAM19}, and astronomy~\cite{DBLP:journals/ascom/VioDA13,scargle1982studies}. Existing time series forecasting methods~\cite{DBLP:conf/icde/LiuMXZLZLZ25, DBLP:conf/iclr/NieNSK23, DBLP:conf/iclr/WuHLZ0L23} mainly focus on fully observed data and cannot adapt to irregularly sampled time series (ISTS)~\cite{DBLP:conf/kdd/0003Y0025, DBLP:conf/icml/ZhangYL0024, mercatali2024graph,DBLP:conf/icml/LuoZ0025}, which are more common in real-world scenarios due to sensor malfunctions, network failure, and varying sampling sources. ISTS exhibits asynchronous observations on non-uniform time intervals across variables~\cite{DBLP:conf/icml/SchirmerELR22}, making it difficult to achieve accurate forecasting for potential informed decision-making~\cite{DBLP:conf/icml/LiL0ZL025}.

Existing ISTS forecasting methods can be generally categorized into three paradigms based on how to model temporal irregularities. The first category focuses on continuous-time modeling~\cite{DBLP:conf/icde/ZhangWYZXBW25,DBLP:conf/iclr/OhLK24}, which often utilizes differential equations or state-space models to naturally handle uneven time gaps. 
The second category of methods employs geometric deep learning~\cite{DBLP:conf/aaai/YalavarthiMSABJ24,DBLP:conf/icml/LiL0ZL025} or patch-based modeling~\cite{DBLP:conf/icml/ZhangYL0024,DBLP:conf/icml/LuoZ0025} to capture dependencies among asynchronous observations via graph connectivity or segmented temporal tokens, which often involve aggregation or pooling operations that may obscure fine-grained temporal patterns. Recently, another line of methods has emerged that applies pre-trained language models (PLMs)~\cite{DBLP:conf/kdd/0003Y0025} for ISTS forecasting due to the generalization capabilities of PLMs. 

Despite these advancements, most existing methods remain confined to a single modality and rely solely on historical observations. These methods often overlook the rich semantic information and fine-grained temporal patterns. Recent studies demonstrate that additional modalities, such as text and images, are capable of providing complementary information to facilitate time series modeling~\cite{DBLP:conf/icml/ZhongR0LW025, DBLP:conf/iclr/0005WMCZSCLLPW24}. Specifically, the textual data often contains contextual descriptions and dataset statistics, which can enhance the understanding of time series patterns. Thus, prompt-based methods~\cite{DBLP:conf/icde/LiuMXZLZLZ25, DBLP:conf/iclr/0005WMCZSCLLPW24} emerge by mapping time series into prompts to help the LLMs understand the time series in depth. These methods often focus on addressing the modality gap between continuous time series and discrete text, aiming to alleviate information misalignment~\cite{DBLP:conf/aaai/0003X0YZ00025}. However, these methods struggle to capture fine-grained temporal patterns, i.e., the ability to learn subtle dynamics, which are particularly important for ISTS forecasting to alleviate the influence of irregularity. More recent studies address this problem by converting time series into their visual versions, such as line graphs or gray-scale images, enabling spatial pattern capturing, which is embedded in time series~\cite{chenvisionts}. Nonetheless, these vision-based methods fall short in learning contextual semantics, since they fail to incorporate domain-specific knowledge.

As a result, we need a new kind of method that can bridge the temporal observations, textual data, and images.
However, it is non-trivial to develop this kind of model, due
to the following challenges.
Although multimodal LLMs (MLLMs) offer a promising means to bridge this gap, leveraging their general understanding capabilities, it is challenging to utilize MLLMs \textcolor{black}{for ISTS forecasting}.
First, a significant representational discrepancy exists between sparse ISTS and the dense inputs required by MLLMs. Naive conversion methods, such as converting time series into standard images or plain text, may not capture the critical uneven time intervals or learn temporal patterns with missing observations. For instance, standard image resizing may distort temporal scales, while linear text serialization may lose the structural correlation across variables.
Second, it is challenging to alleviate the modality gap by \textcolor{black}{aligning temporal observations, visual inputs, and textual context}.
Irregular numerical observations often require high precision, whereas MLLMs operate on a coarse-grained semantic level~\cite{DBLP:conf/icml/ZhongR0LW025}. \textcolor{black}{An effective mechanism is needed to align such threefold representations.} 

\enlargethispage{2\baselineskip}
To address these problems, we propose \emph{MM-ISTS}, which utilizes multimodal vision-text LLMs for ISTS forecasting.
MM-ISTS consists of four major components: the Cross-Modal Vision-Text Encoding module, the ISTS Encoding module, the Adaptive Query-Based Feature Extractor, and the Multimodal Alignment.
Specifically, MM-ISTS employs a Cross-Modal Vision-Text Encoding module to automatically transform ISTS \textcolor{black}{into visual and textual modalities}, which can effectively preserve the irregularity. To better understand temporal correlations, we convert the original ISTS into 3-channel images, with channels for raw observational values, missingness masks, and temporal intervals, enabling the MLLMs to distinguish missing data. Further, we generate descriptive textual prompts with statistics (e.g., missing rates and variable ranges) and dataset-specific domain knowledge to provide complementary contextual information for MLLM feature extraction.
Moreover, to capture temporal dynamics, we propose a customized ISTS Encoding module in parallel with the Vision-Text Encoding branch, which uses a multi-view embedding mechanism to project asynchronous observations, timestamps, and variable indices into a unified latent space. This is realized by a two-stage Transformer encoder that sequentially captures intra-series temporal dependencies and inter-series variable correlations, resulting in robust ISTS representations.


To efficiently align the high-dimensional semantic space of MLLMs with ISTS representations, we propose an Adaptive Query-Based Feature Extractor, which employs a set of learnable queries to extract visual-textual information from MLLMs. This mechanism acts as an information bottleneck, compressing the MLLM output token sequence into fixed-length representations aligned with ISTS representations, filtering out redundant noise while reducing computational overhead during the fusion stage.
Finally, a Multimodal Alignment module is designed to fuse ISTS representations with multimodal representations. It includes a Modality-Aware Gating network that handles irregular data statistics (such as local sparsity levels and variance) to dynamically assign importance weights between the ISTS Encoding module and Cross-Modal Vision-Text Encoding module. This allows the model to use general knowledge from MLLMs when numerical data is sparse, enabling accurate predictions with low-quality data.

Our main contributions are summarized as follows:
\begin{itemize}[leftmargin=0.4cm, topsep=0.4pt]
    \item To the best of our knowledge, we propose MM-ISTS, the first multimodal ISTS forecasting framework augmented by vision-text LLMs, \textcolor{black}{using temporal, visual, and textual modalities}.
    \item We design a Cross-Modal Vision-Text Encoding module that automatically converts ISTS into irregularity-aware images and \textcolor{black}{structured text prompts}, \textcolor{black}{while the ISTS Encoding branch extracts enriched temporal features from historical observations}.
    \item We propose an Adaptive Query-Based Feature Extractor to \textcolor{black}{compress MLLM token embeddings into multimodal representations}, and a \textcolor{black}{Modality-Aware Gating mechanism} to align heterogeneous multimodal features and mitigate the modality gap.
    \item \textcolor{black}{Extensive experiments on real-world datasets demonstrate that MM-ISTS outperforms state-of-the-art baselines}.
\end{itemize}

\section{Related Work}

\textbf{Multi-modal Time Series Forecasting.} Recent studies have incorporated textual information~\cite{DBLP:conf/ijcai/0003ZX000025} for regular time series forecasting to enhance predictive performance beyond numerical signals. For example, Time-LLM~\citep{DBLP:conf/iclr/0005WMCZSCLLPW24} proposes a reprogramming framework that adapts large language models for general time series forecasting.
It aligns time series and textual modalities by converting time series into text prototypes and introducing a Prompt-as-Prefix mechanism to guide the LLM in transforming time series patches for prediction.
TimeCMA~\citep{DBLP:conf/aaai/0003X0YZ00025} proposes a cross-modality alignment framework that aligns disentangled time-series embeddings with robust prompt-based embeddings from large language models.
TimeKD~\citep{DBLP:conf/icde/LiuMXZLZLZ25} introduces a cross-modal LLM-based framework for multivariate time series forecasting via privileged knowledge distillation. It uses a calibrated LLM teacher with privileged information to distill the learned knowledgeable representations into a lightweight student model.
Beyond text and time series modalities, Time-VLM~\citep{DBLP:conf/icml/ZhongR0LW025} integrates an extra vision modality by encoding time series as images and fusing the images, text, and time series with a vision-language model to enhance time series forecasting. Despite recent progress, multimodal large-model approaches still mainly target regularly sampled time series (RSTS), while multimodal ISTS forecasting remains underexplored.

\noindent\textbf{ISTS Forecasting.} 
Existing ISTS forecasting methods~\citep{DBLP:conf/nips/TashiroSSE21,kidger2020neural,DBLP:conf/icml/SchirmerELR22,DBLP:conf/icml/ZhangYL0024,mercatali2024graph,DBLP:conf/icml/LuoZ0025} can be categorized into three categories. The first category focuses on modeling continuous temporal dynamics. Latent ODE~\citep{DBLP:conf/nips/RubanovaCD19} introduces continuous-time hidden state evolution via neural ordinary differential equations, while CRU~\citep{DBLP:conf/icml/SchirmerELR22} adopts a state-space formulation with Kalman filtering. The second category represents ISTS using relational or patch-based structures. GraFITi~\citep{DBLP:conf/aaai/YalavarthiMSABJ24} formulates ISTS as a bipartite graph between variables and timestamps, T-PatchGNN~\citep{DBLP:conf/icml/ZhangYL0024} segments ISTS into temporal patches and models dependencies via Transformers and GNNs, and HyperIMTS~\citep{DBLP:conf/icml/LiL0ZL025} models ISTS with hypergraph structures.
More recently, pre-trained large language models (LLMs) have been explored for ISTS forecasting. ISTS-PLM~\citep{DBLP:conf/kdd/0003Y0025} investigates how representation schemes affect LLMs' ability to model ISTS, and Time-IMM~\citep{chang2026time} introduces a benchmark focusing on realistic irregular sampling. Despite these advances, most existing approaches either concentrate on unimodal ISTS or lack global contextual information to adequately model complex and dynamic data patterns.

\section{Preliminary}

\textbf{Irregularly-Sampled Time Series (ISTS).} We consider an ISTS consisting of $N$ variables. For the $n$-th variable, its historical observations are denoted as
$o_n = \{ (t^n_i, x^n_i) \}_{i=1}^{L_n}$, where $t^n_i \in \mathbb{R}$ denotes the
timestamp of the $i$-th observation and $x^n_i \in \mathbb{R}$ denotes the
corresponding recorded value. The number of observations $L_n$ may vary across
variables, and the complete ISTS is denoted by
$\mathcal{O} = \{ o_n \}_{n=1}^N$.

\begin{figure*}[t]
    \centering
    \includegraphics[width=\linewidth]{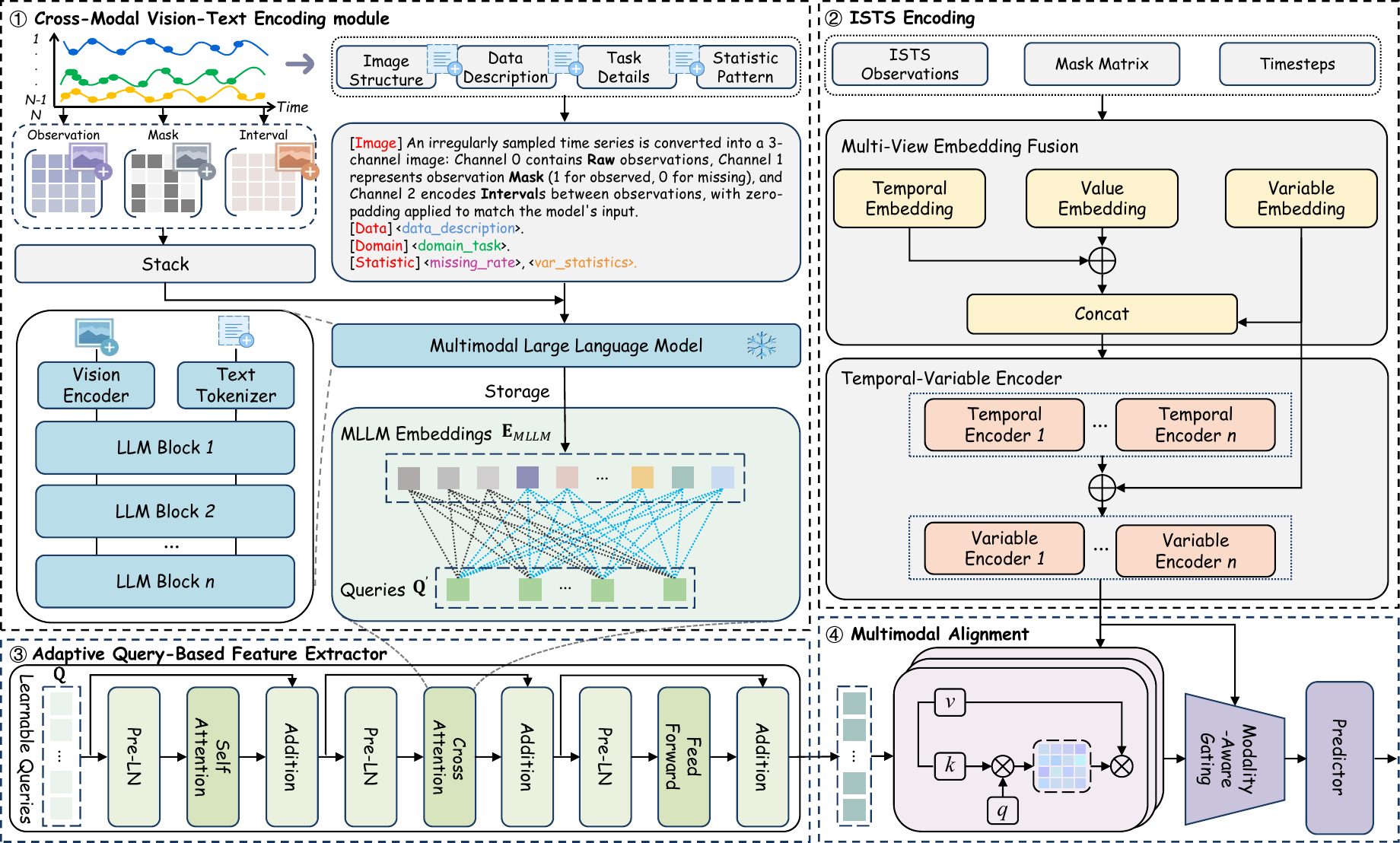}
    \vspace{-0.7cm}
    \caption{Overview of the proposed MM-ISTS framework.}
    \label{fig:framework}
    \vspace{-0.35cm}
\end{figure*}

\noindent\textbf{Canonical Representation.} In practice, a commonly adopted preprocessing strategy is the canonical
pre-alignment representation
~\citep{che2018recurrent,DBLP:conf/iclr/ShuklaM21,DBLP:conf/iclr/ZhangZTZ22,DBLP:conf/kdd/ZhangZCBL23,DBLP:conf/nips/RubanovaCD19,DBLP:conf/nips/BrouwerSAM19,DBLP:conf/nips/BilosSRJG21,DBLP:conf/icml/SchirmerELR22},
which transforms the ISTS data $\mathcal{O}$ into a triplet
$(\mathcal{T}, \mathcal{X}, \mathcal{M})$. Here,
$\mathcal{T} = [t_l]_{l=1}^{L} \in \mathbb{R}^{L}$ denotes the ordered set of unique
timestamps obtained by merging all observation times across variables, i.e.,
$\mathcal{T} = \bigcup_{n=1}^{N} \{ t^n_i \}_{i=1}^{L_n}$.
The value matrix
$\mathcal{X} = [[x^n_l]_{n=1}^N]_{l=1}^L \in \mathbb{R}^{L \times N}$ aligns
multivariate observations on the unified timeline, where $x^n_l$ records the
observed value of the $n$-th variable at timestamp $t_l$ if available, and is
filled with zero otherwise. To explicitly distinguish real observations from
filled values, a binary mask
$\mathcal{M} = [[m^n_l]_{n=1}^N]_{l=1}^L \in \{0,1\}^{L \times N}$ is introduced,
where $m^n_l = 1$ indicates that variable $n$ is observed at time $t_l$, and
$m^n_l = 0$ otherwise. Together, the triplet
$(\mathcal{T}, \mathcal{X}, \mathcal{M})$ provides a fixed-shape representation on
a shared temporal grid while preserving the original irregular sampling pattern
through the observation mask.
To better preserve the native irregular sampling characteristics of each variable, we refer to the representation strategy in ISTS-PLM~\citep{DBLP:conf/kdd/0003Y0025} and adopt a variable-independent sequence representation for ISTS. Specifically, we do not merge timestamps across variables; instead, each variable maintains its own ordered timestamps, denoted as $\mathcal{T}_n = [t^n_i]_{i=1}^{L_n} \in \mathbb{R}^{L_n}$ for the $n$-th variable, where $L_n$ is the number of observations of the $n$-th variable and $t^n_i$ represents the $i$-th observation timestamp of the $n$-th variable. 

\noindent\textbf{Irregularly-Sampled Time Series Forecasting.} Given an ISTS $\mathcal{O}$, we define
forecasting queries to specify which future values are to be predicted. For the $n$-th variable, a forecasting query is defined by a future timestamp
$q^n_j$ satisfying $q^n_j > \max_i t^n_i$, where $j = 1, \dots, Q_n$ denotes the
query timestamps for variable $n$. The set of forecasting queries is denoted as
$\mathcal{Q} = \{ \{ q^n_j \}_{j=1}^{Q_n} \}_{n=1}^N$.

The goal of ISTS forecasting is to learn a model $F_\theta$, parameterized by $\theta$, that maps historical observations and forecasting queries to future value predictions.
\begin{equation}
F_\theta(\mathcal{O}, \mathcal{Q})
\;\longrightarrow\;
\widehat{\mathcal{X}}
=
\bigl\{
  \{ \hat{x}^n_j \}_{j=1}^{Q_n}
\bigr\}_{n=1}^N,
\end{equation}
where $\hat{x}^n_j$ denotes the predicted value of the $n$-th variable at the $j$-th query timestamp $q^n_j$.
\section{Methodology}

We present \textit{MM-ISTS}, a multimodal framework designed to tackle the challenges of ISTS forecasting. The core idea of MM-ISTS is to combine the precise numerical patterns learned from historical data with multimodal representations extracted by pre-trained MLLMs. As illustrated in Figure~\ref{fig:framework}, the framework comprises four components: (1) a Cross-Modal Vision-Text Encoding module that transforms ISTS into irregularity-aware visual and textual representations; (2) a Dual-stage ISTS Encoding branch that sequentially models intra-series temporal dynamics and inter-series variable correlations; (3) an Adaptive Query-Based Feature Extractor that converts MLLM tokens into variable-aligned query embeddings; and (4) a Multimodal Alignment module equipped with a Modality-Aware Gating mechanism for adaptive fusion.

\subsection{Cross-Modal Vision-Text Encoding}
\label{sec:vision_text}

Frozen pretrained MLLMs provide strong feature extraction and cross-modal understanding abilities. To leverage these abilities for ISTS forecasting, we first transform each irregular sample into visual and textual inputs while preserving values, missingness, and unequal time intervals. This conversion allows the MLLM to extract useful multimodal features from the irregular observation patterns.

\subsubsection{Irregularity-Aware Image Construction}

Traditional time-series-to-image methods, such as line plots, connect adjacent observations and therefore introduce interpolation bias between irregular samples, potentially distorting the original observation pattern. In contrast, we use observed values, observation masks, and temporal intervals to construct a three-channel irregularity-aware image as the visual input for the MLLM.

Given an ISTS sample with $N$ variables and maximum history length $L$, we construct a three-channel image $\mathcal{I} \in \mathbb{R}^{3 \times N \times L}$. The image height corresponds to the variable dimension, and the width follows the ordered historical positions of each variable.

\textbf{Observed Data} channel $\mathbf{C}_0 \in \mathbb{R}^{N \times L}$ stores the measured values. For variable $n$, $(\mathbf{C}_0)_{n,l}$ records the value at the $l$-th historical position when valid. For invalid or padded positions, $(\mathbf{C}_0)_{n,l}$ is set to $0$ and the validity is indicated by the mask channel.

\textbf{Missingness Mask} channel $\mathbf{C}_1 \in \mathbb{R}^{N \times L}$ records whether each position is valid. Specifically, $(\mathbf{C}_1)_{n,l}=1$ indicates that the corresponding position contains a valid observation, while $(\mathbf{C}_1)_{n,l}=0$ indicates an invalid or padded position.

\textbf{Temporal Interval} channel $\mathbf{C}_2 \in \mathbb{R}^{N \times L}$ encodes irregular sampling intervals. Since variables may be observed at different timestamps, the time gap is computed independently for each variable. For variable $n$ at the $l$-th valid observation, we define $\delta^n_l = t^n_l - t^n_{l-1}$, with $\delta^n_1 = 0$ for the first valid observation, and set $(\mathbf{C}_2)_{n,l}=\delta^n_l$ for valid positions. For invalid or padded positions, $(\mathbf{C}_2)_{n,l}$ is set to $0$ and the mask channel distinguishes them from valid intervals.

The final irregularity-aware image $\mathcal{I}$ is obtained by stacking these three channels:
\begin{equation}
    \mathcal{I} = \mathit{Stack}(\mathbf{C}_0, \mathbf{C}_1, \mathbf{C}_2) \in \mathbb{R}^{3 \times N \times L}.
\end{equation}
This construction represents values, observation validity, and local time intervals separately within the same sample, allowing the MLLM to process irregular observation information through a visual input. Before feeding into the MLLM, the image is resized to match the expected input resolution and normalized to the appropriate pixel value range. Appendix~\ref{sec:image_input_scale} reports the resulting image sizes and preprocessing details before MLLM input.

\subsubsection{Structured Text Prompting}

We construct a Text Prompt Template $\mathcal{P}$ with four components: image construction description, data description, forecasting task description, and statistics of observed variables. The image input describes the irregular sample in a visual form, while the text input adds context about image construction, dataset background, forecasting objective, and variable statistics for each sample. This design provides the frozen MLLM with aligned visual and textual context for extracting multimodal representations in ISTS forecasting.

We compute three summary statistics for each variable $n$: the observed mean $\mu_n = \frac{\sum_{l=1}^{L} m^n_l x^n_l}{\sum_{l=1}^{L} m^n_l}$, the observed range $[x_n^{\min}, x_n^{\max}]$, and the missing rate $\rho_n = 1 - \frac{1}{L}\sum_{l=1}^{L} m^n_l$. These statistics give a compact description of scale and sparsity for each variable. To avoid unreliable summaries for extremely sparse variables, we set $\mathcal{S}_n = \mathit{Format}(\mu_n, x_n^{\min}, x_n^{\max})$ only when $\rho_n \le \tau$; otherwise, $\mathcal{S}_n = \emptyset$.

The final Text Prompt Template $\mathcal{P}$ is assembled by concatenating four types of instruction components:
\begin{equation}
    \mathcal{P} = [\mathcal{P}_{\mathit{img}}, \mathcal{P}_{\mathit{data}}, \mathcal{P}_{\mathit{task}}, \{\mathcal{S}_n\}_{n: \rho_n \le \tau}],
\end{equation}
where $\mathcal{P}_{\mathit{img}}$ explains the construction of the three-channel image, $\mathcal{P}_{\mathit{data}}$ provides a concise description summarized from dataset background information, $\mathcal{P}_{\mathit{task}}$ specifies the forecasting task, and $\{\mathcal{S}_n\}$ provides statistics of observed variables computed from the historical input sequence.

\subsubsection{MLLM Feature Extraction}

The frozen MLLM encoder $\mathcal{E}_{\mathit{MLLM}}$ jointly processes the image $\mathcal{I}$ and text prompt $\mathcal{P}$, producing visual-textual hidden states without updating pretrained parameters.

We use hidden states from deep MLLM layers after image-text token interaction:
\begin{equation}
    \mathbf{E}_{\mathit{MLLM}} = \mathcal{E}_{\mathit{MLLM}}(\mathcal{I}, \mathcal{P}) \in \mathbb{R}^{S \times d_m},
\end{equation}
where $S$ is the number of MLLM token embeddings and $d_m$ is the MLLM hidden dimension. The MLLM parameters remain frozen during training, which preserves the pre-trained knowledge. Following TimeCMA~\citep{DBLP:conf/aaai/0003X0YZ00025}, we pre-compute and store the MLLM token embeddings before training to improve computational efficiency.

\subsection{ISTS Encoding}
\label{sec:ists_encoding}

While MLLMs provide high-level visual and textual representations, they may not capture fine-grained numerical patterns as accurately as dedicated time series models. Our ISTS encoder addresses this limitation by operating on carefully designed embeddings followed by a Transformer encoder that models temporal dynamics within each variable and correlations across variables.

\subsubsection{Multi-View Embedding Fusion}
In order to better capture the relationships within ISTS using Transformer architectures, we model them from different perspectives.

\textbf{Temporal Embedding.} Unlike discrete positional encodings used in standard Transformers, ISTS requires embeddings that can handle continuous and irregularly spaced timestamps. We employ a learnable sinusoidal mapping $\phi: \mathbb{R} \to \mathbb{R}^D$ that captures both periodic patterns and linear temporal trends:
\begin{equation}
    \phi(t)_d = 
    \begin{cases} 
    \omega_0 t + \beta_0, & d = 0, \\ 
    \sin(\omega_d t + \beta_d), & d > 0,
    \end{cases}
\end{equation}
where $\{\omega_d\}_{d=0}^{D-1}$ and $\{\beta_d\}_{d=0}^{D-1}$ are learnable parameters. The first dimension captures linear time progression, while the remaining dimensions encode periodic patterns at different frequencies.

\textbf{Variable Embedding.} To distinguish between different variables and enable the model to learn variable-specific patterns, we introduce a learnable embedding matrix $\mathbf{E}_{\mathit{var}} \in \mathbb{R}^{N \times D}$ that assigns a unique representation $\mathbf{e}_n^{\mathit{var}} \in \mathbb{R}^D$ to each variable index $n \in \{1, \dots, N\}$. These embeddings are learned during training and capture the intrinsic characteristics of each variable type.

\textbf{Value Embedding.} For each observation, we need to encode both the numerical value and whether it was actually observed. We concatenate the observed value $x^n_l$ with its corresponding mask indicator $m^n_l$ and apply a linear projection:
$\mathbf{e}_{l,n}^{\mathit{val}} = [x^n_l, m^n_l] \mathbf{W}_{\mathit{val}} + \mathbf{b}_{\mathit{val}}$,
where $\mathbf{W}_{\mathit{val}} \in \mathbb{R}^{2 \times D}$ and $\mathbf{b}_{\mathit{val}} \in \mathbb{R}^D$ are learnable parameters. 

\textbf{Embedding Fusion.} The fused embedding for the $l$-th time step of variable $n$ combines temporal and value information through a mask-gated mechanism: $\mathbf{z}_{l,n} = m^n_l \cdot \phi(t^n_l) + \mathbf{e}_{l,n}^{\mathit{val}}$. The mask-gated design ensures that temporal information is weighted by observation presence. To enable the model to aggregate information at the variable level, we prepend the variable embedding $\mathbf{e}_n^{\mathit{var}}$ as a learnable prompt token to each variable's sequence, forming $\mathbf{Z}_n = [\mathbf{e}_n^{\mathit{var}}, \mathbf{z}_{1,n}, \dots, \mathbf{z}_{L,n}] \in \mathbb{R}^{(L+1) \times D}$.

\vspace{-0.15\baselineskip}
\subsubsection{Temporal-Variable Encoder}

Multivariate ISTS exhibit two types of dependencies: temporal dependencies within each variable and cross-variable dependencies. To disentangle and effectively model these two types of relationships, we employ a Temporal Encoder and a Variable Encoder using stacked Transformer layers.

\textbf{Temporal Encoder.} We apply a multi-layer Transformer encoder $\mathcal{F}_{\mathit{temp}}$ independently to each variable's sequence to capture temporal patterns. The encoder consists of $L_t$ stacked layers, where each layer applies multi-head self-attention followed by a feed-forward network. The attention mechanism allows each position to attend to all other positions within the same variable's sequence:
\begin{equation}
    \mathit{Attention}(\mathbf{Q}, \mathbf{K}, \mathbf{V}) = \mathit{Softmax}\left(\frac{\mathbf{Q}\mathbf{K}^\top}{\sqrt{d_k}}\right)\mathbf{V},
\end{equation}
where $\mathbf{Q}, \mathbf{K}, \mathbf{V} \in \mathbb{R}^{(L+1) \times d_k}$ are the query, key, and value matrices obtained by linear projections, and $d_k$ is the dimension per attention head. The multi-head attention extends this by computing $h$ parallel attention heads and concatenating their outputs:
\begin{equation}
    \mathit{MultiHead}(\mathbf{Z}) = \mathit{Concat}(\mathit{head}_1, \dots, \mathit{head}_h)\mathbf{W}^O,
\end{equation}
where $\mathit{head}_i = \mathit{Attention}(\mathbf{Z}\mathbf{W}^Q_i, \mathbf{Z}\mathbf{W}^K_i, \mathbf{Z}\mathbf{W}^V_i)$ and $\mathbf{W}^O$ is the output projection matrix.

The Temporal Encoder processes each variable's sequence independently, producing $\mathbf{H}_n^{\mathit{temp}} = \mathcal{F}_{\mathit{temp}}(\mathbf{Z}_n) \in \mathbb{R}^{(L+1) \times D}$. Since different variables may have different numbers of observed time points, we perform mask-aware aggregation to obtain a fixed-length representation $\mathbf{h}_n \in \mathbb{R}^D$ for each variable:
\begin{equation}
    \mathbf{h}_n = \frac{\sum_{l=0}^{L} \tilde{m}^n_l \cdot \mathbf{H}_{n,l}^{\mathit{temp}}}{\sum_{l=0}^{L} \tilde{m}^n_l},
\end{equation}
where $\tilde{m}^n_0 = 1$ for variable tokens and $\tilde{m}^n_l = m^n_l$ for $l \geq 1$.

\textbf{Variable Encoder.} We then model the dependencies across different variables. The aggregated variable representations are stacked to form a matrix $\mathbf{H}_{\mathit{agg}} = [\mathbf{h}_1, \dots, \mathbf{h}_N]^\top \in \mathbb{R}^{N \times D}$. We enhance these representations by adding the variable embeddings and apply a multi-layer Transformer encoder $\mathcal{F}_{\mathit{var}}$ consisting of $L_v$ layers to model cross-variable correlations. The Variable Encoder allows each variable's representation to attend to and incorporate information from all other variables. The final output is $\mathbf{H}_{\mathit{ISTS}} = \mathcal{F}_{\mathit{var}}(\mathbf{H}_{\mathit{agg}} + \mathbf{E}_{\mathit{var}}) \in \mathbb{R}^{N \times D}$.

\subsection{Adaptive Query-Based Feature Extractor}
\label{sec:qformer}

While the MLLM output $\mathbf{E}_{\mathit{MLLM}} \in \mathbb{R}^{S \times d_m}$ encodes contextual information from visual and textual modalities, it cannot be directly fused with the ISTS Encoding branch output because different image and text inputs produce MLLM token sequences with different lengths, making them difficult to align with the temporal features learned by the ISTS Encoding branch. To bridge this representational gap, we propose an Adaptive Query-Based Feature Extractor inspired by the Q-Former architecture~\citep{DBLP:conf/icml/0008LSH23}, which acts as a learnable information bottleneck that compresses and aligns the sequence of MLLM token embeddings into variable-aligned representations.

We introduce a set of $N$ learnable query tokens $\mathbf{Q} \in \mathbb{R}^{N \times d_m}$, with one query associated with each variable. The queries interact with the MLLM output $\mathbf{E}_{\mathit{MLLM}}$ through $K$ stacked layers, each containing query self-attention, cross-attention to MLLM features, and a feed-forward network. This design lets learnable queries exchange information and extract relevant visual-textual information from the full hidden-state sequence.

For each layer, we first apply layer normalization and self-attention to the query tokens:
\begin{equation}
    \tilde{\mathbf{Q}} = \mathbf{Q} + \mathit{MultiHead}_{\mathit{self}}(\mathit{LN}(\mathbf{Q})),
\end{equation}
where $\mathit{MultiHead}_{\mathit{self}}$ denotes multi-head self-attention with queries, keys, and values all derived from the input.

We set $\mathbf{Q}' = \mathit{LN}(\tilde{\mathbf{Q}})$ and compute projected queries $\mathbf{Q}_p = \mathbf{Q}' \mathbf{W}^Q$, keys $\mathbf{K}_p = \mathbf{E}_{\mathit{MLLM}} \mathbf{W}^K$, values $\mathbf{V}_p = \mathbf{E}_{\mathit{MLLM}} \mathbf{W}^V$, where $\mathbf{W}^Q, \mathbf{W}^K, \mathbf{W}^V$ are learnable projection matrices. The cross-attention and the residual connection are:
\begin{equation}
    \mathit{CrossAttn}(\mathbf{Q}', \mathbf{E}_{\mathit{MLLM}}, \mathbf{E}_{\mathit{MLLM}}) = \mathit{Softmax}\left(\frac{\mathbf{Q}_p \mathbf{K}_p^\top}{\sqrt{d_k}}\right) \mathbf{V}_p,
\end{equation}
\begin{equation}
    \hat{\mathbf{Q}} = \tilde{\mathbf{Q}} + \mathit{MultiHead}_{\mathit{cross}}(\mathbf{Q}', \mathbf{E}_{\mathit{MLLM}}, \mathbf{E}_{\mathit{MLLM}}).
\end{equation}

Finally, a feed-forward network produces $\mathbf{Q}'' = \hat{\mathbf{Q}} + \mathit{FFN}(\mathit{LN}(\hat{\mathbf{Q}}))$.
The output $\mathbf{Q}''$ serves as the query input to the next layer.

\setcounter{dbltopnumber}{1}
\begin{table*}[!t]
\centering
\caption{Overall performance comparison on four datasets. The best results are highlighted in bold, and the second-best are underlined.}
\vspace{-0.3cm}
\label{tab:main_results}
\resizebox{\linewidth}{!}{
\begin{tabular}{l|c c|c c|c c|c c}
\toprule
\textbf{Dataset}
& \multicolumn{2}{c|}{\textbf{PhysioNet}}
& \multicolumn{2}{c|}{\textbf{MIMIC}}
& \multicolumn{2}{c|}{\textbf{Human Activity}}
& \multicolumn{2}{c}{\textbf{USHCN}} \\
\midrule
\textbf{Metric}
& MSE$\times 10^{-3}$ & MAE$\times 10^{-2}$
& MSE$\times 10^{-2}$ & MAE$\times 10^{-2}$
& MSE$\times 10^{-3}$ & MAE$\times 10^{-2}$
& MSE$\times 10^{-1}$ & MAE$\times 10^{-1}$ \\
\midrule
DLinear & 41.86 $\pm$ 0.05 & 15.52 $\pm$ 0.03 & 4.90 $\pm$ 0.00 & 16.29 $\pm$ 0.05 & 4.03 $\pm$ 0.01 & 4.21 $\pm$ 0.01 & 6.21 $\pm$ 0.00 & 3.88 $\pm$ 0.02 \\
TimesNet & 16.48 $\pm$ 0.11 & 6.14 $\pm$ 0.03 & 5.88 $\pm$ 0.08 & 13.62 $\pm$ 0.07 & 3.12 $\pm$ 0.01 & 3.56 $\pm$ 0.02 & 5.58 $\pm$ 0.05 & 3.60 $\pm$ 0.04 \\
PatchTST & 12.00 $\pm$ 0.23 & 6.02 $\pm$ 0.14 & 3.78 $\pm$ 0.03 & 12.43 $\pm$ 0.10 & 4.29 $\pm$ 0.14 & 4.80 $\pm$ 0.09 & 5.75 $\pm$ 0.01 & 3.57 $\pm$ 0.02 \\
Crossformer & 6.66 $\pm$ 0.11 & 4.81 $\pm$ 0.11 & 2.65 $\pm$ 0.10 & 9.56 $\pm$ 0.29 & 4.29 $\pm$ 0.20 & 4.89 $\pm$ 0.17 & 5.25 $\pm$ 0.04 & 3.27 $\pm$ 0.09 \\
Graph WaveNet & 6.04 $\pm$ 0.28 & 4.41 $\pm$ 0.11 & 2.93 $\pm$ 0.09 & 10.50 $\pm$ 0.15 & 2.89 $\pm$ 0.03 & 3.40 $\pm$ 0.05 & 5.29 $\pm$ 0.04 & 3.16 $\pm$ 0.09 \\
MTGNN & 6.26 $\pm$ 0.18 & 4.46 $\pm$ 0.07 & 2.71 $\pm$ 0.23 & 9.55 $\pm$ 0.65 & 3.03 $\pm$ 0.03 & 3.53 $\pm$ 0.03 & 5.39 $\pm$ 0.05 & 3.34 $\pm$ 0.02 \\
StemGNN & 6.86 $\pm$ 0.28 & 4.76 $\pm$ 0.19 & 1.73 $\pm$ 0.02 & 7.71 $\pm$ 0.11 & 8.81 $\pm$ 0.37 & 6.90 $\pm$ 0.02 & 5.75 $\pm$ 0.09 & 3.40 $\pm$ 0.09 \\
CrossGNN & 7.22 $\pm$ 0.36 & 4.96 $\pm$ 0.12 & 2.95 $\pm$ 0.16 & 10.82 $\pm$ 0.21 & 3.03 $\pm$ 0.10 & 3.48 $\pm$ 0.08 & 5.66 $\pm$ 0.04 & 3.53 $\pm$ 0.05 \\
FourierGNN & 6.84 $\pm$ 0.35 & 4.65 $\pm$ 0.12 & 2.55 $\pm$ 0.03 & 10.22 $\pm$ 0.08 & 2.99 $\pm$ 0.02 & 3.42 $\pm$ 0.02 & 5.82 $\pm$ 0.06 & 3.62 $\pm$ 0.07 \\
\midrule
GRU-D & 5.59 $\pm$ 0.09 & 4.08 $\pm$ 0.05 & 1.76 $\pm$ 0.03 & 7.53 $\pm$ 0.09 & 2.94 $\pm$ 0.05 & 3.53 $\pm$ 0.06 & 5.54 $\pm$ 0.38 & 3.40 $\pm$ 0.28 \\
SeFT & 9.22 $\pm$ 0.18 & 5.40 $\pm$ 0.08 & 1.87 $\pm$ 0.01 & 7.84 $\pm$ 0.08 & 12.20 $\pm$ 0.17 & 8.43 $\pm$ 0.07 & 5.80 $\pm$ 0.19 & 3.70 $\pm$ 0.11 \\
RainDrop & 9.82 $\pm$ 0.08 & 5.57 $\pm$ 0.06 & 1.99 $\pm$ 0.03 & 8.27 $\pm$ 0.07 & 14.92 $\pm$ 0.14 & 9.45 $\pm$ 0.05 & 5.78 $\pm$ 0.22 & 3.67 $\pm$ 0.17 \\
Warpformer & 5.94 $\pm$ 0.35 & 4.21 $\pm$ 0.12 & 1.73 $\pm$ 0.04 & 7.58 $\pm$ 0.13 & 2.79 $\pm$ 0.04 & 3.39 $\pm$ 0.03 & 5.25 $\pm$ 0.05 & 3.23 $\pm$ 0.05 \\
mTAND & 6.23 $\pm$ 0.24 & 4.51 $\pm$ 0.17 & 1.85 $\pm$ 0.06 & 7.73 $\pm$ 0.13 & 3.22 $\pm$ 0.07 & 3.81 $\pm$ 0.07 & 5.33 $\pm$ 0.05 & 3.26 $\pm$ 0.10 \\
\midrule
Latent-ODE & 6.05 $\pm$ 0.57 & 4.23 $\pm$ 0.26 & 1.89 $\pm$ 0.19 & 8.11 $\pm$ 0.52 & 3.34 $\pm$ 0.11 & 3.94 $\pm$ 0.12 & 5.62 $\pm$ 0.03 & 3.60 $\pm$ 0.12 \\
CRU & 8.56 $\pm$ 0.26 & 5.16 $\pm$ 0.09 & 1.97 $\pm$ 0.02 & 7.93 $\pm$ 0.19 & 6.97 $\pm$ 0.78 & 6.30 $\pm$ 0.47 & 6.09 $\pm$ 0.17 & 3.54 $\pm$ 0.18 \\
Neural Flow & 7.20 $\pm$ 0.07 & 4.67 $\pm$ 0.04 & 1.87 $\pm$ 0.05 & 8.03 $\pm$ 0.19 & 4.05 $\pm$ 0.13 & 4.46 $\pm$ 0.09 & 5.35 $\pm$ 0.05 & 3.25 $\pm$ 0.05 \\
T-PatchGNN & \underline{5.11 $\pm$ 0.11} & 3.73 $\pm$ 0.11 & 1.66 $\pm$ 0.02 & 7.21 $\pm$ 0.14 & 2.79 $\pm$ 0.11 & 3.24 $\pm$ 0.07 & \textbf{5.03 $\pm$ 0.04} & 3.14 $\pm$ 0.09 \\
KAFNet & 5.47 $\pm$ 0.07 & 3.82 $\pm$ 0.12 & 1.70 $\pm$ 0.02 & 7.23 $\pm$ 0.04 & 2.70 $\pm$ 0.05 & \underline{3.16 $\pm$ 0.03} & 5.19 $\pm$ 0.14 & 3.17 $\pm$ 0.10 \\
APN & 5.40 $\pm$ 0.05 & 3.68 $\pm$ 0.07 & \underline{1.66 $\pm$ 0.01} & \underline{7.01 $\pm$ 0.03} & 2.83 $\pm$ 0.01 & 3.21 $\pm$ 0.02 & 5.44 $\pm$ 0.08 & 3.08 $\pm$ 0.07 \\
\midrule
Time-LLM & 10.41 $\pm$ 0.80 & 6.27 $\pm$ 0.29 & 2.35 $\pm$ 0.06 & 9.86 $\pm$ 0.38 & 4.98 $\pm$ 0.03 & 4.67 $\pm$ 0.01 & 6.16 $\pm$ 0.07 & 4.01 $\pm$ 0.09 \\
TimeCMA & 7.56 $\pm$ 1.83 & 4.95 $\pm$ 0.29 & 1.99 $\pm$ 0.21 & 7.38 $\pm$ 0.30 & 3.43 $\pm$ 0.05 & 3.81 $\pm$ 0.04 & 5.85 $\pm$ 0.09 & 3.81 $\pm$ 0.12 \\
Time-VLM & 24.22 $\pm$ 1.16 & 8.24 $\pm$ 0.29 & 6.81 $\pm$ 0.16 & 15.98 $\pm$ 0.22 & 4.94 $\pm$ 0.09 & 4.68 $\pm$ 0.07 & 6.08 $\pm$ 0.11 & 4.12 $\pm$ 0.05 \\
ISTS-PLM & 5.17 $\pm$ 0.13 & \underline{3.67 $\pm$ 0.06} & 1.72 $\pm$ 0.05 & 7.18 $\pm$ 0.34 & \underline{2.61 $\pm$ 0.07} & 3.19 $\pm$ 0.07 & 5.28 $\pm$ 0.05 & \underline{3.03 $\pm$ 0.06} \\
\midrule
MM-ISTS (Ours) &
\textbf{4.98 $\pm$ 0.11} &
\textbf{3.54 $\pm$ 0.05} &
\textbf{1.63 $\pm$ 0.02} &
\textbf{6.85 $\pm$ 0.07} &
\textbf{2.47 $\pm$ 0.01} &
\textbf{3.06 $\pm$ 0.02} &
\underline{5.10 $\pm$ 0.03} &
\textbf{2.94 $\pm$ 0.05} \\
\bottomrule
\end{tabular}
}
\end{table*}

After $K$ layers, the query tokens summarize relevant contextual information from the visual-textual hidden states and form the multimodal representation $\mathbf{H}_{\mathit{MM}} = \mathbf{Q}'' \in \mathbb{R}^{N \times d_m}$. Each output is associated with a variable, giving the Vision-Text Encoding branch a variable-wise organization consistent with the ISTS Encoding branch before fusion.

\subsection{Multimodal Alignment}
\label{sec:alignment}

To integrate precise numerical patterns with contextual knowledge, we propose an alignment module that adaptively fuses $\mathbf{H}_{\mathit{ISTS}} \in \mathbb{R}^{N \times D}$ and $\mathbf{H}_{\mathit{MM}} \in \mathbb{R}^{N \times d_m}$ based on variable-specific data quality.

\textbf{Cross-Attention Fusion.} Direct concatenation or addition of the two representations cannot explicitly capture the interactions between numerical ISTS features and MLLM-derived features. Instead, we use cross-attention to enable the numerical features to selectively query and incorporate relevant contextual information:
\begin{equation}
    \mathbf{H}_{\mathit{fused}} = \mathit{CrossAttn}(\mathbf{H}_{\mathit{ISTS}}, \mathbf{H}_{\mathit{MM}}, \mathbf{H}_{\mathit{MM}}) \in \mathbb{R}^{N \times D},
\end{equation}
where $\mathbf{H}_{\mathit{ISTS}}$ serves as the query and $\mathbf{H}_{\mathit{MM}}$ serves as keys and values. 

\textbf{Modality-Aware Gating.} Different variables in an ISTS may have different observation densities. For a densely observed variable, the numerical features from the ISTS encoder are reliable and should receive larger weights. Conversely, for a sparsely observed variable with high missing rates, the contextual information from MLLMs may provide more valuable information. To address such variable-specific differences in observation quality, we introduce a Modality-Aware Gating mechanism that adaptively balances the contributions of the two modalities for each variable.

For each variable $n$, we compute a statistics vector $\mathbf{s}_n \in \mathbb{R}^{4}$ that summarizes its data characteristics:
\begin{equation}
    \mathbf{s}_n = [\mu_n, \sigma_n, \rho_n, c_n],
\end{equation}
where $\mu_n$ is the mean of observed values, $\sigma_n$ is the standard deviation, $\rho_n$ is the missing rate, and $c_n = \sum_{l=1}^{L} m^n_l / L$ is the normalized observation count. 

A gating network $\mathcal{G}$, implemented as a two-layer MLP with ReLU activation, maps this statistics vector to fusion weights:
\begin{equation}
    [\alpha_n^{\mathit{num}}, \alpha_n^{\mathit{mm}}] = \mathit{Softmax}(\mathcal{G}(\mathbf{s}_n)) \in \mathbb{R}^2,
\end{equation}
where $\alpha_n^{\mathit{num}} + \alpha_n^{\mathit{mm}} = 1$.

The final fused representation for each variable is computed using the Modality-Aware Gating weights:
\begin{equation}
    \mathbf{H}_{\mathit{final}}[n] = \alpha_n^{\mathit{num}} \cdot \mathbf{H}_{\mathit{ISTS}}[n] + \alpha_n^{\mathit{mm}} \cdot \mathbf{H}_{\mathit{fused}}[n].
\end{equation}

\subsection{ISTS Predictor}

Given the fused representation $\mathbf{H}_{\mathit{final}}$ and forecasting queries $\mathcal{Q} = \{ \{ q^n_j \}_{j=1}^{Q_n} \}_{n=1}^N$, we generate predictions by conditioning variable features on target timestamps. For query $q^n_j$, the prediction is generated via an MLP:
\begin{equation}
    \hat{x}^n_j = \mathit{MLP}([\mathbf{H}_{\mathit{final}}[n] \,\|\, q^n_j]).
\end{equation}
The trainable modules are optimized via MSE loss over all valid queries, with the MLLM backbone frozen:
\begin{equation}
    \mathcal{L} = \frac{1}{\sum_{n=1}^{N} Q_n} \sum_{n=1}^{N} \sum_{j=1}^{Q_n} (\hat{x}^n_j - x^n_j)^2.
\end{equation}

\section{Experiments}
\begingroup
\raggedbottom
\setlength{\textfloatsep}{6pt plus 1pt minus 1pt}
\setlength{\intextsep}{6pt plus 1pt minus 1pt}
\setlength{\floatsep}{6pt plus 1pt minus 1pt}
\setlength{\abovecaptionskip}{4pt}
\setlength{\belowcaptionskip}{0pt}
\subsection{Experimental Setup}
\subsubsection{Datasets}
To comprehensively evaluate the effectiveness of the proposed method on ISTS forecasting, we conduct experiments on four widely used benchmark datasets: \textit{PhysioNet}~\citep{silva2012predicting}, \textit{MIMIC}~\citep{johnson2016mimic}, \textit{Human Activity}~\citep{vidulin2010localization}, and \textit{USHCN}~\citep{menne2015long}. These datasets span diverse application domains, including healthcare, biomechanics, and climate science. We randomly split each dataset into training, validation, and test sets with a ratio of $6{:}2{:}2$.

\vspace{-2pt}

\subsubsection{Baselines}
To evaluate MM-ISTS against representative alternatives, we compare it with baselines from four groups:
\textit{(1) Regular Time Series Forecasting Models:} DLinear~\citep{DBLP:conf/aaai/ZengCZ023}, TimesNet~\citep{DBLP:conf/iclr/WuHLZ0L23}, PatchTST~\citep{DBLP:conf/iclr/NieNSK23}, Crossformer~\citep{DBLP:conf/iclr/ZhangY23}, Graph WaveNet~\citep{DBLP:conf/ijcai/WuPLJZ19}, MTGNN~\citep{DBLP:conf/kdd/WuPL0CZ20}, StemGNN~\citep{DBLP:conf/nips/CaoWDZZHTXBTZ20}, CrossGNN~\citep{DBLP:conf/nips/HuangSZDWZW23}, and FourierGNN~\citep{yi2023fouriergnn}.
\textit{(2) ISTS Classification and Imputation Models:} GRU-D~\citep{che2018recurrent}, SeFT~\citep{DBLP:conf/icml/HornMBRB20}, RainDrop~\citep{DBLP:conf/iclr/ZhangZTZ22}, Warpformer~\citep{DBLP:conf/kdd/ZhangZCBL23}, and mTAND~\citep{DBLP:conf/iclr/ShuklaM21}.
\textit{(3) ISTS Forecasting Models:} Latent ODEs~\citep{DBLP:conf/nips/RubanovaCD19}, CRU~\citep{DBLP:conf/icml/SchirmerELR22}, Neural Flows~\citep{DBLP:conf/nips/BilosSRJG21}, T-PatchGNN~\citep{DBLP:conf/icml/ZhangYL0024}, KAFNet~\citep{DBLP:journals/corr/abs-2508-01971}, and APN~\citep{liu2026apn}.
\textit{(4) Large-Model-Based Time Series Models:} Time-LLM~\citep{DBLP:conf/iclr/0005WMCZSCLLPW24}, TimeCMA~\citep{DBLP:conf/aaai/0003X0YZ00025}, Time-VLM~\citep{DBLP:conf/icml/ZhongR0LW025}, and ISTS-PLM~\citep{DBLP:conf/kdd/0003Y0025}. This group includes methods that use pretrained LLMs or VLMs for time-series modeling. Details of the baseline methods and experimental implementation are provided in Appendix~\ref{sec:baseline_details} and Appendix~\ref{sec:implementation_details}.

\enlargethispage{0.8\baselineskip}
\vspace{-3pt}
\subsection{Experimental Results}

\subsubsection{Main Results}
Table~\ref{tab:main_results} presents the forecasting performance of MM-ISTS and all baseline methods across four ISTS datasets.
MM-ISTS achieves the best overall performance. It ranks first on both MSE and MAE for PhysioNet, MIMIC, and Human Activity, and obtains the best MAE and second-best MSE on USHCN. Compared with the average performance of the specialized ISTS forecasting baselines, namely Latent ODEs, CRU, Neural Flows, T-PatchGNN, KAFNet, and APN, our MM-ISTS reduces MSE by \textbf{17.8\%} and MAE by \textbf{15.3\%} across the four datasets. Compared with the multimodal time-series forecasting baselines Time-LLM, TimeCMA, and Time-VLM, which are mainly designed for regular time series, it reduces MSE by \textbf{45.2\%} and MAE by \textbf{35.0\%} on average. Compared with ISTS-PLM, the strongest baseline using a pretrained LLM, MM-ISTS reduces MSE by \textbf{4.4\%} and MAE by \textbf{3.8\%} on average across the four datasets.
The baseline groups reveal complementary limitations: Regular Time Series Forecasting Models are designed for regular time series and may not capture irregular sampling characteristics; classification and imputation models are not directly optimized for future forecasting objectives; specialized ISTS forecasters mainly use numerical histories; and large model baselines either focus on regular time series or omit multimodal information. MM-ISTS addresses these gaps by effectively combining numerical ISTS representations with multimodal representations extracted from irregularity-aware images and structured text prompts, and then aligning these representations for forecasting. Its gains on clinical, sensor, and climate datasets suggest that the advantage remains visible across domains and sampling densities.

\subsubsection{Ablation Study}
\begin{figure}[t]
  \centering
  \begin{subfigure}[b]{0.48\linewidth}
    \centering
    \includegraphics[width=\linewidth]{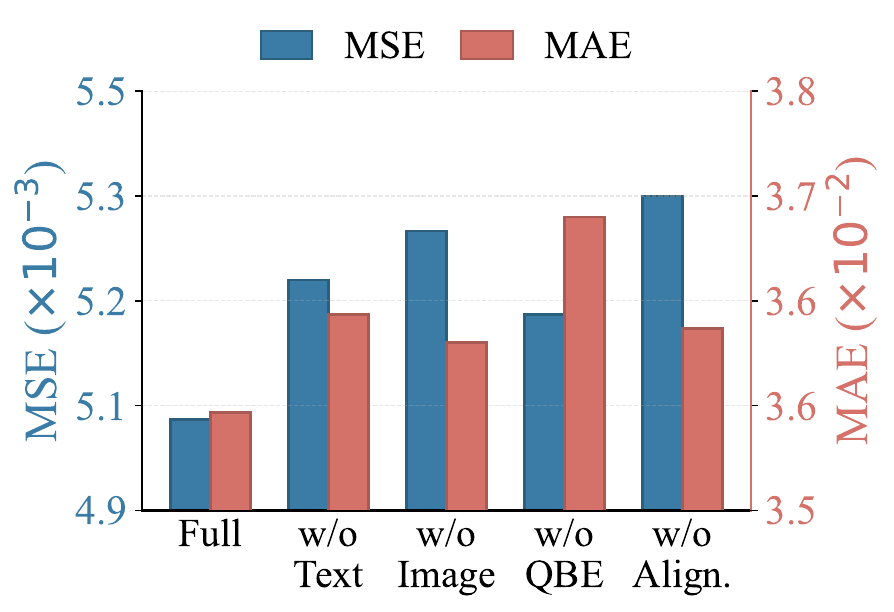}
    \caption{PhysioNet}
  \end{subfigure}
  \hfill
  \begin{subfigure}[b]{0.48\linewidth}
    \centering
    \includegraphics[width=\linewidth]{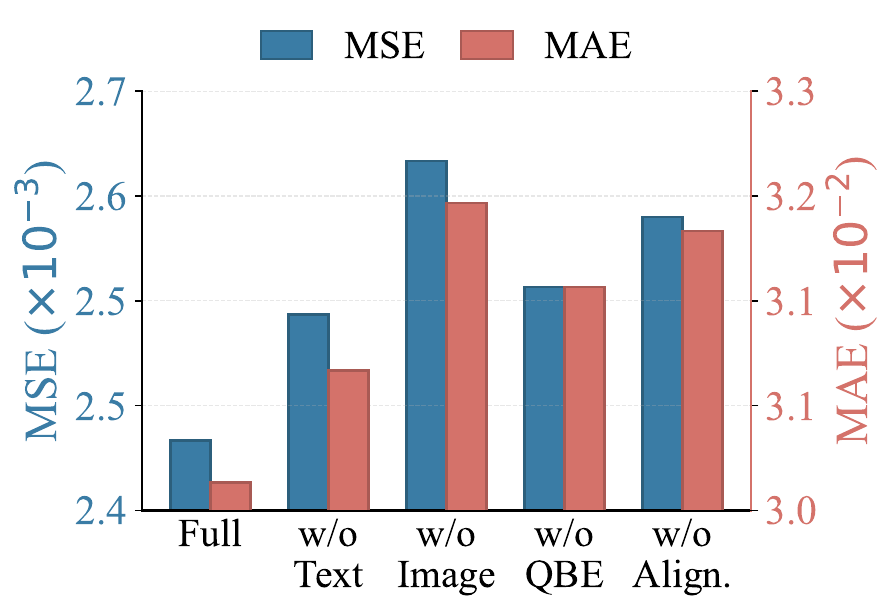}
    \caption{Human Activity}
  \end{subfigure}
  \vspace{-4pt}
  \caption{Ablation study.}
  \vspace{-4pt}
  \label{fig:ablation1}
\end{figure}

We evaluate the contribution of major components in MM-ISTS on PhysioNet and Human Activity. We compare the full model with four variants: \textit{w/o Text}, which removes the Text Prompt Template $\mathcal{P}$; \textit{w/o Image}, which excludes the irregularity-aware image representation $\mathcal{I}$; \textit{w/o QBE}, which replaces the proposed Adaptive Query-Based Feature Extractor (QBE) with the final MLLM token embedding; and \textit{w/o Align}, which substitutes the Multimodal Alignment module with element-wise addition. As shown in Figure~\ref{fig:ablation1}, removing any component leads to performance degradation on both datasets. For \textit{w/o Text}, the consistent performance gap indicates that the Text Prompt Template $\mathcal{P}$ provides useful complementary information. For \textit{w/o Image}, the observed decline shows that the irregularity-aware image helps the model use observation values, missingness, and time intervals through the MLLM branch. For \textit{w/o QBE}, the degradation suggests that a single MLLM token embedding cannot provide sufficiently variable-aligned multimodal information. For \textit{w/o Align}, the performance drop shows that multimodal information needs explicit alignment before it is fused with the ISTS Encoding branch.

\begin{figure}[!htbp]
  \centering
  \begin{subfigure}[t]{0.47\linewidth}
    \centering
    \includegraphics[width=\linewidth]{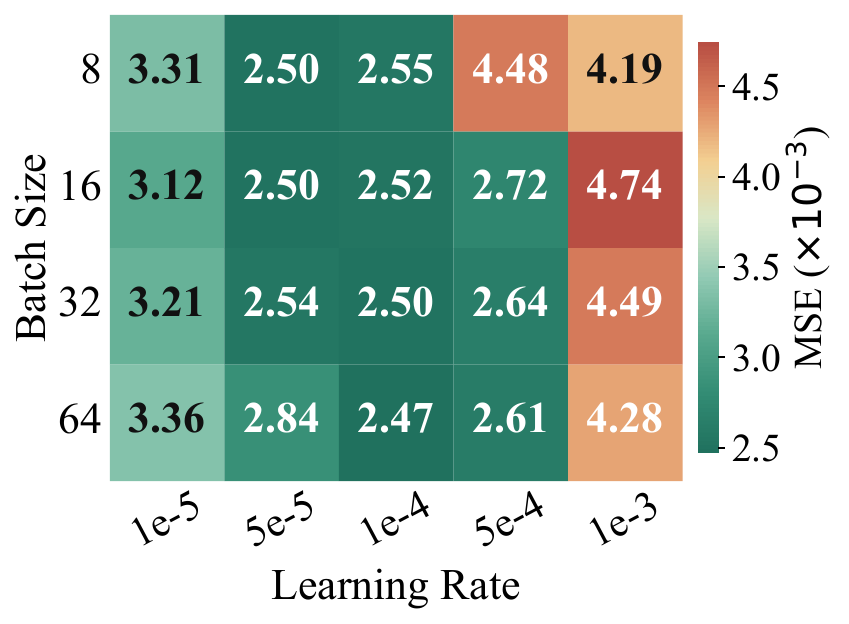}
    \caption{Parameter sensitivity}
    \label{fig:lr_bs_heatmap}
  \end{subfigure}
  \hspace{0.04\linewidth}
  \begin{subfigure}[t]{0.47\linewidth}
    \centering
    \includegraphics[width=0.9\linewidth]{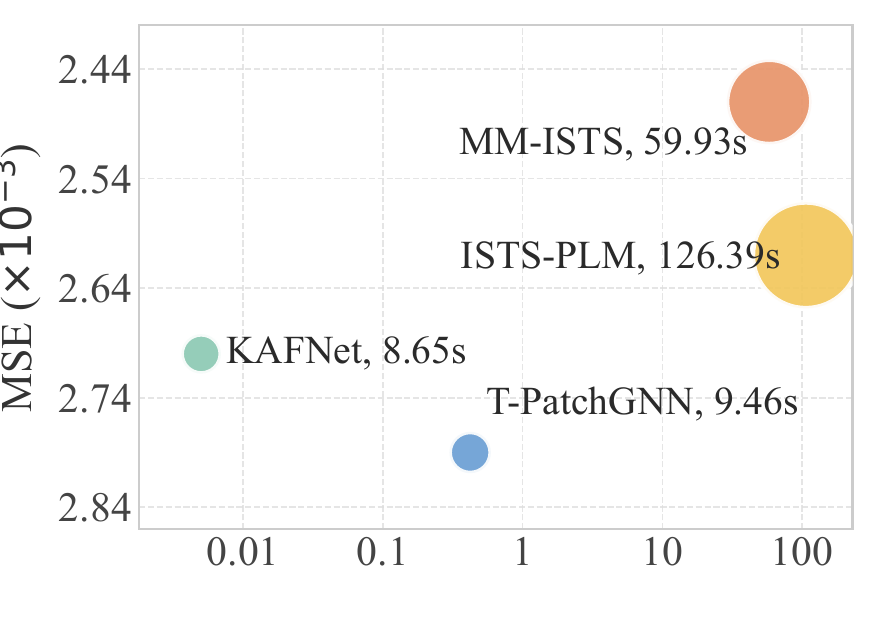}
    \caption{Efficiency}
    \label{fig:efficiency_comparison}
  \end{subfigure}
  \vspace{-4pt}
  \caption{Parameter sensitivity and efficiency analysis.}
  \label{fig:human_activity_efficiency_heatmap}
  \vspace{-4pt}
\end{figure}

\begin{figure*}[!t]
  \centering
  \begin{subfigure}[b]{0.31\textwidth}
    \centering
    \includegraphics[width=\linewidth]{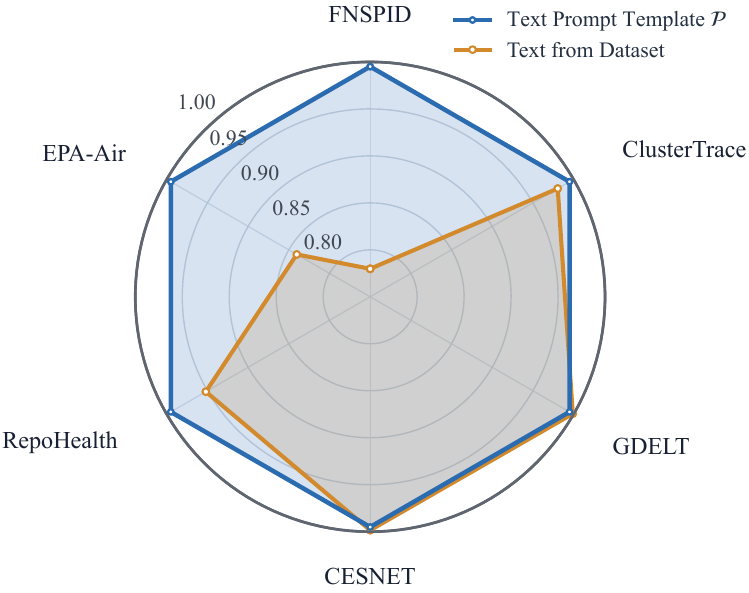}
    \caption{Text source comparison}
    \label{fig:text_prompt_source}
  \end{subfigure}
  \hfill
  \begin{subfigure}[b]{0.31\textwidth}
    \centering
    \includegraphics[width=\linewidth]{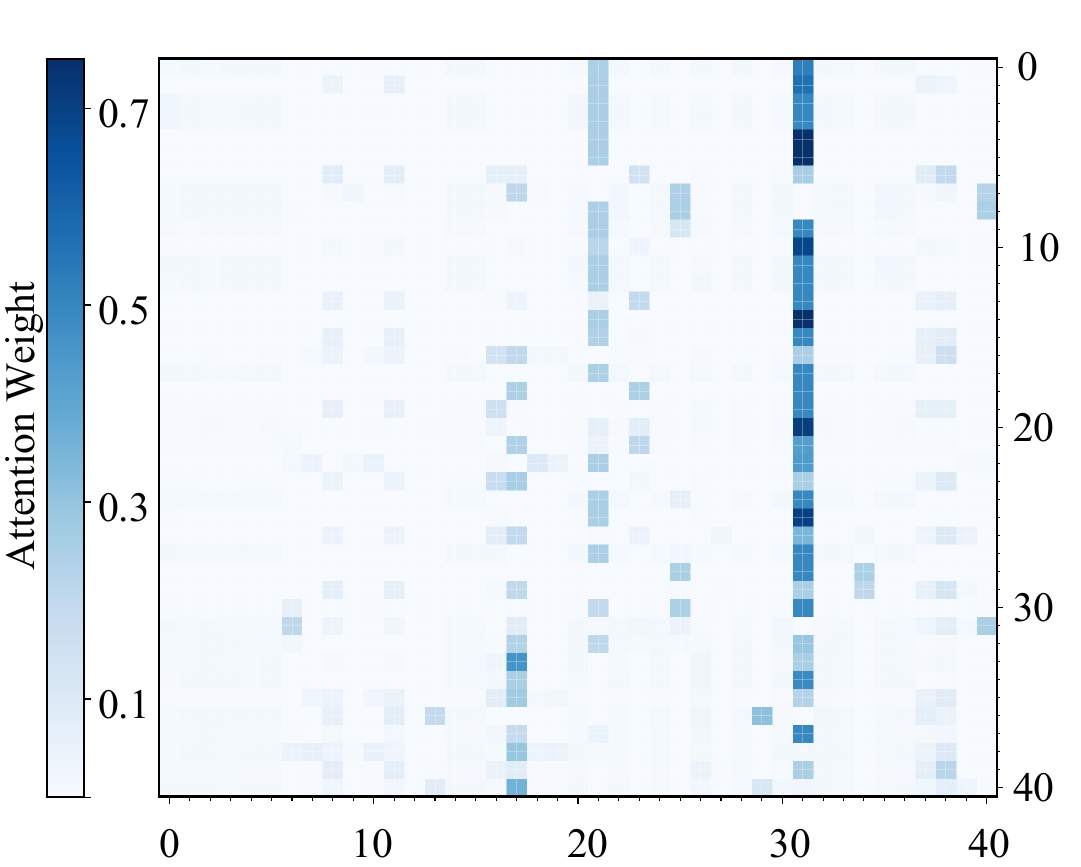}
    \caption{Attention map}
    \label{fig:case_attention}
  \end{subfigure}
  \hfill
  \begin{subfigure}[b]{0.31\textwidth}
    \centering
    \includegraphics[width=\linewidth]{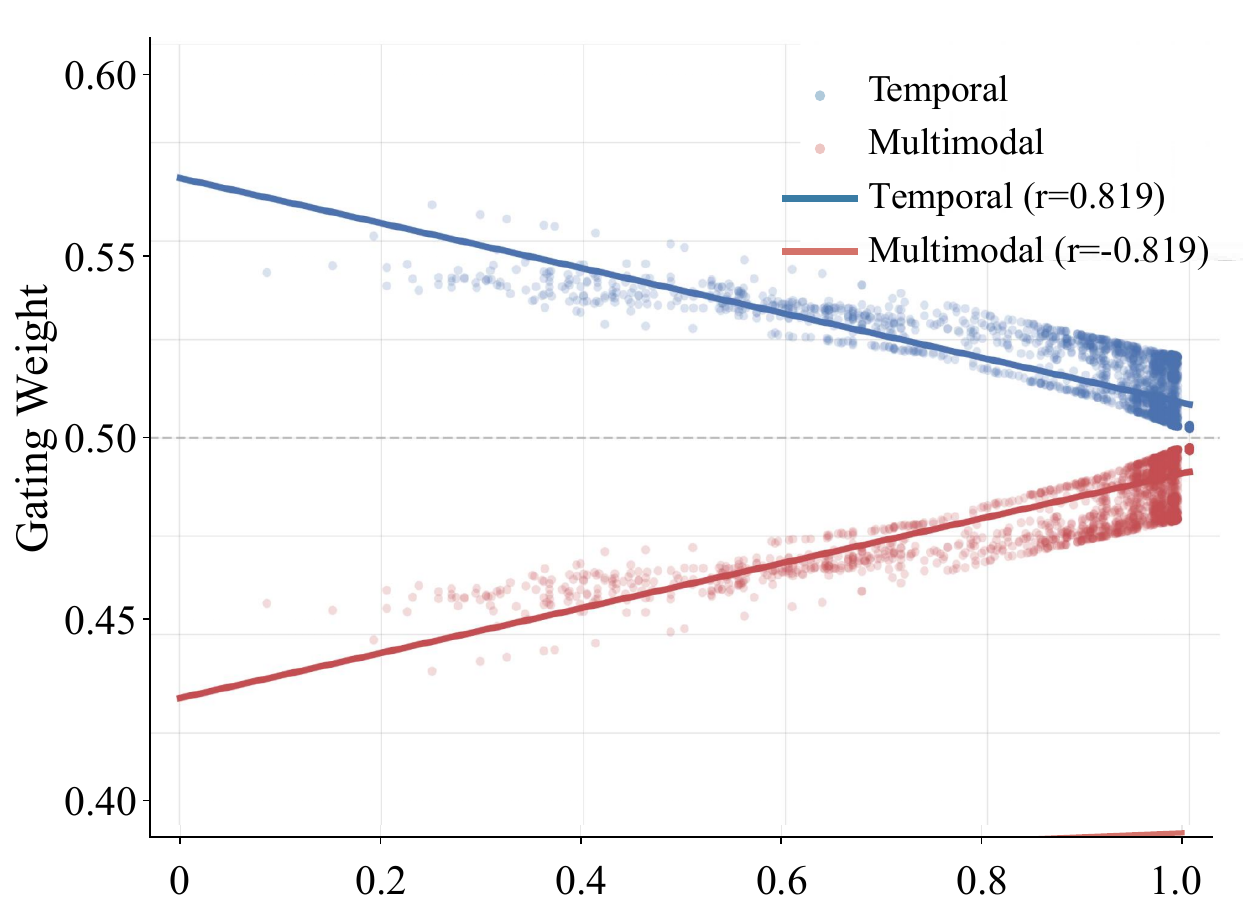}
    \caption{Gating weights vs. sparsity}
    \label{fig:case_gating}
  \end{subfigure}
  \vspace{-2pt}
  \caption{Text source comparison, cross-attention patterns, and adaptive gating behavior.}
  \label{fig:text_case_combined}
  \vspace{-4pt}
\end{figure*}

\vspace{-2pt}
\subsubsection{Text Source Analysis}
\label{sec:text_source_main}
Time-IMM~\citep{chang2026time} contains irregular multivariate time series paired with timestamped text summaries generated from domain-specific documents, reports, or logs. We use six Time-IMM datasets covering financial markets, cluster workloads, global events, network traffic, repository activity, and air-quality monitoring. We compare our Structured Text Prompting strategy with the original texts included in the Time-IMM multimodal datasets. Figure~\ref{fig:text_case_combined}~(a) reports normalized radar scores across these datasets, where larger values indicate lower MSE. The Text Prompt Template $\mathcal{P}$ consistently improves MSE across all six datasets, with larger gains on FNSPID and EPA-Air and smaller but stable gains on ClusterTrace, GDELT, and CESNET. For the Time-IMM datasets, our Structured Text Prompting strategy constructs text prompts with image construction, data description, forecasting task, and statistics of observed variables. Compared with using the Time-IMM text, our Text Prompt Template $\mathcal{P}$ provides more comprehensive task-related information. This result indicates that combining these components provides more useful textual context for forecasting.

\subsubsection{Parameter Sensitivity}
\label{sec:parameter_sensitivity_main}
Figure~\ref{fig:human_activity_efficiency_heatmap}~(a) summarizes how different learning rate and batch size settings interact on Human Activity. Learning rate has a more visible effect on performance than batch size. Moderate learning rates, especially $5\times10^{-5}$ and $10^{-4}$, give lower errors across most batch sizes, whereas the largest learning rate leads to higher errors. The best region appears around medium learning rates and relatively large batch sizes, which is consistent with the configuration used in the main experiments. The heatmap also indicates that changing the batch size within this region causes smaller variation than moving to an overly large learning rate. Additional sensitivity analyses over model depth, hidden dimension, backbone choice, and PhysioNet settings are reported in Appendix~\ref{sec:hyperparameter_appendix}.

\subsubsection{Efficiency Analysis}
\label{sec:efficiency_main}
Figure~\ref{fig:human_activity_efficiency_heatmap}~(b) reports training cost and forecasting error on Human Activity for MM-ISTS, two lightweight forecasting models, KAFNet and T-PatchGNN, and the LLM-based ISTS-PLM. Since both MM-ISTS and ISTS-PLM rely on large language models, the most direct efficiency comparison is between them. ISTS-PLM fine-tunes the language model during training, whereas MM-ISTS freezes the MLLM backbone and only optimizes lightweight downstream modules. In addition, the Adaptive Query-Based Feature Extractor compresses variable-length MLLM hidden states into $N$ variable-aligned query embeddings, so the Multimodal Alignment module avoids processing the full token sequence. As a result, MM-ISTS uses about half the time per epoch of ISTS-PLM and has fewer trainable parameters. KAFNet and T-PatchGNN remain faster as specialized small models. Thus, MM-ISTS is a more efficient LLM-based alternative to methods based on fine-tuning, while achieving lower prediction error than lightweight ISTS models. The PhysioNet result is provided in Appendix~\ref{sec:physionet_efficiency_appendix}.

\subsubsection{Case Study}
Figure~\ref{fig:text_case_combined}~(b) visualizes the attention map in the Multimodal Alignment module, where ISTS features attend to informative multimodal tokens. The attention weights are not uniformly distributed over all tokens; instead, only a subset receives relatively large weights, suggesting selective use of multimodal information. Figure~\ref{fig:text_case_combined}~(c) shows the relationship between modality-aware gating weights and variable sparsity. Variables with high missing rates receive higher weights for multimodal features, while densely observed variables rely more on temporal features. Together, these observations support our design that MLLM outputs provide complementary information when numerical observations are scarce.

\begin{figure}[!htbp]
  \centering
  \captionsetup[subfigure]{font=footnotesize,justification=centering}
  \begin{subfigure}[t]{0.48\linewidth}
    \centering
    \includegraphics[width=\linewidth]{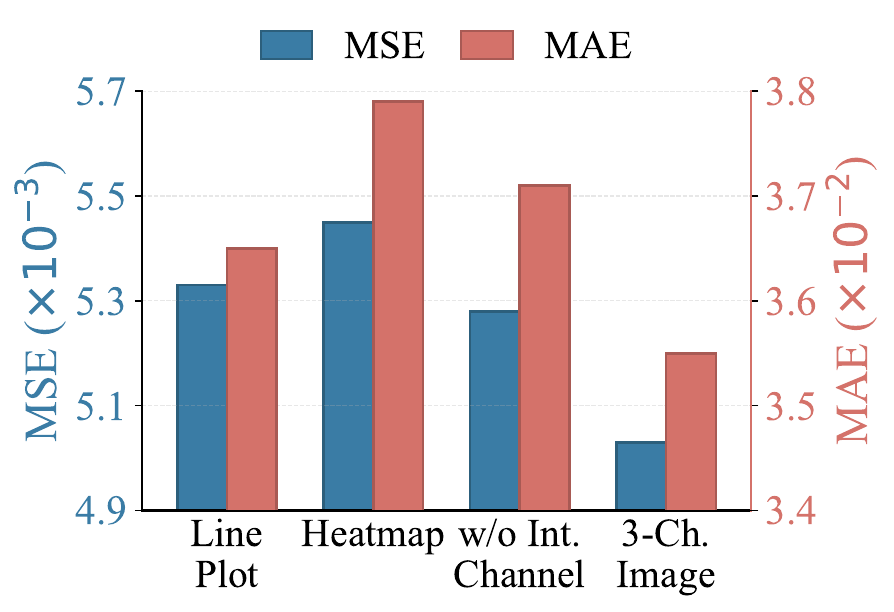}
    \caption{Image variants (PhysioNet)}
  \end{subfigure}
  \hfill
  \begin{subfigure}[t]{0.48\linewidth}
    \centering
    \includegraphics[width=\linewidth]{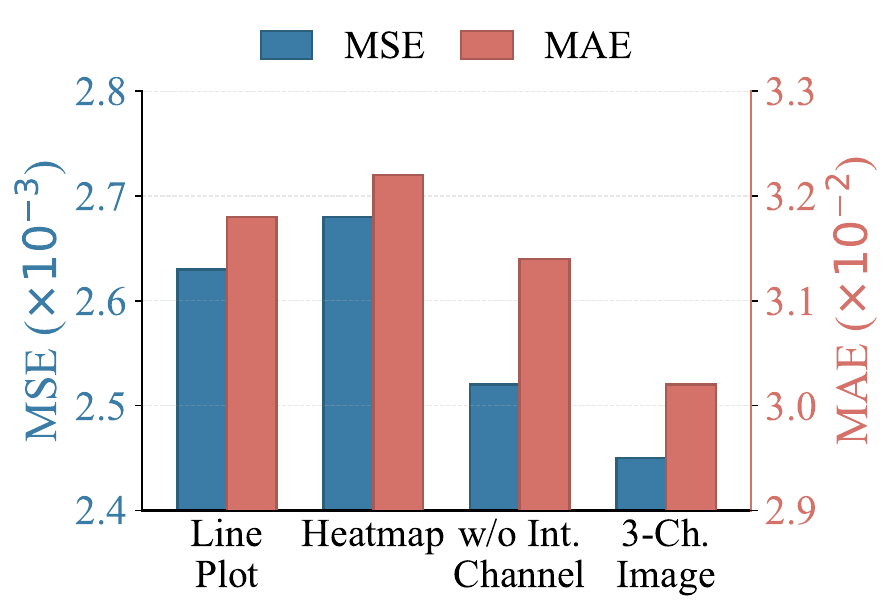}
    \caption{Image variants (Human Activity)}
  \end{subfigure}

  \vspace{1mm}

\begin{subfigure}[t]{0.48\linewidth}
    \centering
    \includegraphics[width=\linewidth]{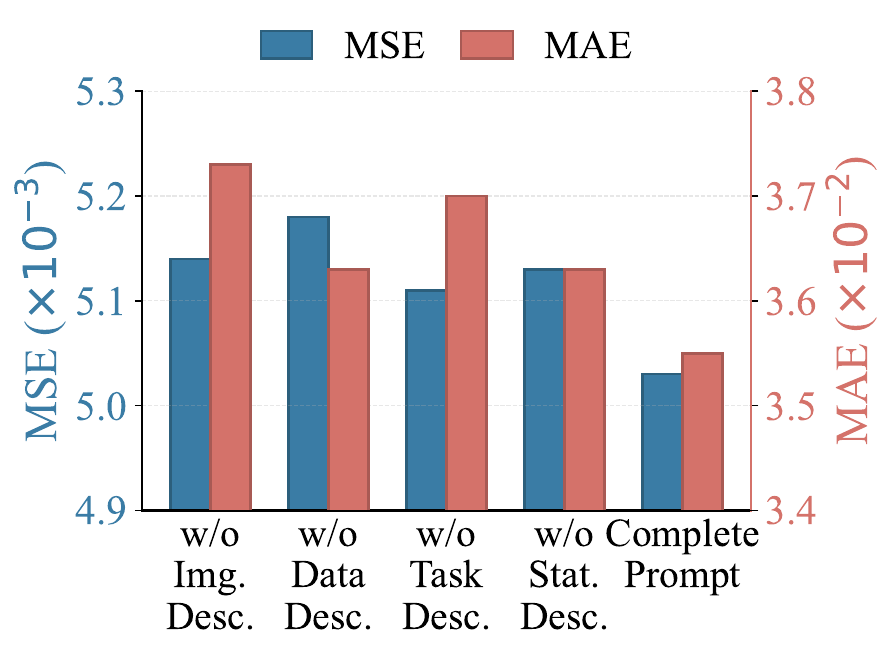}
    \caption{Prompt variants (PhysioNet)}
  \end{subfigure}
  \hfill
  \begin{subfigure}[t]{0.48\linewidth}
    \centering
    \includegraphics[width=\linewidth]{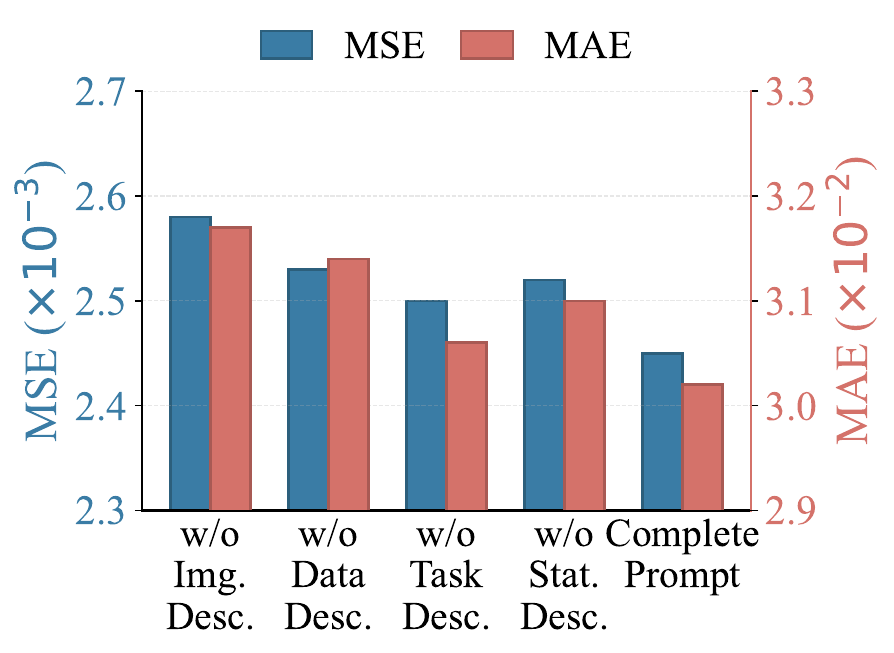}
    \caption{Prompt variants (Human Activity)}
  \end{subfigure}
  \caption{Effect of image construction variants and different prompt components.}
  \label{fig:image_prompt_analysis}
  \vspace{-2pt}
\end{figure}

\subsubsection{Image Construction Analysis}
\label{sec:image_construction_main}
We further study how different image constructions affect the multimodal branch. Figure~\ref{fig:image_prompt_analysis}~(a)--(b) show traditional line plots, heatmaps, a three-channel variant without the temporal interval channel, and our irregularity-aware three-channel image. The proposed design obtains the lowest errors on both PhysioNet and Human Activity. Line plots connect observed points. Heatmaps arrange the value matrix over variables and time steps. These two representations do not explicitly separate observation masks or temporal intervals into dedicated channels. The \textit{w/o Int. Channel} variant constructs the image only from observed values, so it does not encode observation masks or temporal intervals. In contrast, our image keeps observed values, observation masks, and temporal intervals in separate channels for the same ISTS sample. This comparison isolates the effect of explicitly preserving validity and time-gap information beyond value visualization. The results show that, within this controlled comparison, preserving these irregular-sampling signals in the visual input improves forecasting performance on both datasets.

\subsubsection{Prompt Component Analysis}
\label{sec:prompt_component_main}
We analyze prompt components by removing the image construction description, data description, forecasting task description, and statistics of observed variables, respectively. Figure~\ref{fig:image_prompt_analysis}~(c)--(d) shows that the complete prompt performs best on PhysioNet and Human Activity. The image construction description helps the MLLM interpret the three-channel visual input, while the data description and forecasting task description provide the dataset background and define the forecasting task. The statistics component further supplies information from the current input window, such as value ranges, means, and sparsity. Removing any component worsens performance, suggesting that the text prompt template benefits from jointly describing the three-channel image construction, dataset background, prediction task, and statistics of observed variables.
\par\endgroup

\section{Conclusion}
We present MM-ISTS, a multimodal forecasting framework for ISTS. To overcome representational and modality gaps in ISTS, we introduce a cross-modal encoding strategy that transforms sparse ISTS into irregularity-aware visual and textual representations. By combining ISTS representations with a Temporal-Variable Encoder and pretrained MLLMs, MM-ISTS captures both numerical dynamics and complementary semantic information. Furthermore, an Adaptive Query-Based Feature Extractor compresses the sequence of MLLM output token embeddings into variable-aligned multimodal representations. Finally, the Multimodal Alignment module aligns multimodal representations with variable-level temporal representations using cross-attention and a Modality-Aware Gating mechanism. Experiments on real-world benchmarks demonstrate consistent performance improvements over state-of-the-art ISTS forecasting methods and recent LLM-based baselines, highlighting the potential of multimodal learning for ISTS forecasting.

\begingroup
\setlength{\bibsep}{0pt}
\bibliographystyle{ACM-Reference-Format}
\bibliography{sample-base}

@inproceedings{DBLP:conf/cinc/ReynaJSJSWSNC19,
  author       = {Matthew A. Reyna and
                  Christopher Josef and
                  Salman Seyedi and
                  Russell Jeter and
                  Supreeth P. Shashikumar and
                  M. Brandon Westover and
                  Ashish Sharma and
                  Shamim Nemati and
                  Gari D. Clifford},
  title        = {Early Prediction of Sepsis from Clinical Data: the PhysioNet/Computing
                  in Cardiology Challenge 2019},
  booktitle    = {CinC},
  pages        = {1--4},
  year         = {2019}
}

@article{kidger2020neural,
  title={Neural controlled differential equations for irregular time series},
  author={Kidger, Patrick and Morrill, James and Foster, James and Lyons, Terry},
  journal={NeurIPS},
  volume={33},
  pages={6696--6707},
  year={2020}
}

@article{DBLP:journals/ascom/VioDA13,
  author       = {Roberto Vio and
                  Mar{\'{\i}}a D{\'{\i}}az{-}Trigo and
                  Paola Andreani},
  title        = {Irregular time series in astronomy and the use of the Lomb-Scargle
                  periodogram},
  journal      = {Astron. Comput.},
  volume       = {1},
  pages        = {5--16},
  year         = {2013}
}

@inproceedings{DBLP:conf/nips/RubanovaCD19,
  author       = {Yulia Rubanova and
                  Tian Qi Chen and
                  David Duvenaud},
  title        = {Latent Ordinary Differential Equations for Irregularly-Sampled Time
                  Series},
  booktitle    = {NeurIPS},
  pages        = {5321--5331},
  year         = {2019}
}

@inproceedings{DBLP:conf/icml/SchirmerELR22,
  author       = {Mona Schirmer and
                  Mazin Eltayeb and
                  Stefan Lessmann and
                  Maja Rudolph},
  title        = {Modeling Irregular Time Series with Continuous Recurrent Units},
  booktitle    = {ICML},
  pages        = {19388--19405},
  year         = {2022}
}

@inproceedings{DBLP:conf/nips/BilosSRJG21,
  author       = {Marin Bilos and
                  Johanna Sommer and
                  Syama Sundar Rangapuram and
                  Tim Januschowski and
                  Stephan G{\"{u}}nnemann},
  title        = {Neural Flows: Efficient Alternative to Neural ODEs},
  booktitle    = {NeurIPS},
  pages        = {21325--21337},
  year         = {2021}
}

@inproceedings{DBLP:conf/icml/ZhangYL0024,
  author       = {Weijia Zhang and
                  Chenlong Yin and
                  Hao Liu and
                  Xiaofang Zhou and
                  Hui Xiong},
  title        = {Irregular Multivariate Time Series Forecasting: {A} Transformable
                  Patching Graph Neural Networks Approach},
  booktitle    = {ICML},
  year         = {2024}
}

@inproceedings{DBLP:conf/icml/LuoZ0025,
  author       = {Yicheng Luo and
                  Bowen Zhang and
                  Zhen Liu and
                  Qianli Ma},
  title        = {Hi-Patch: Hierarchical Patch {GNN} for Irregular Multivariate Time
                  Series},
  booktitle    = {ICML},
  year         = {2025}
}

@inproceedings{DBLP:conf/kdd/0003Y0025,
  author       = {Weijia Zhang and
                  Chenlong Yin and
                  Hao Liu and
                  Hui Xiong},
  title        = {Unleashing The Power of Pre-Trained Language Models for Irregularly
                  Sampled Time Series},
  booktitle    = {SIGKDD},
  pages        = {3831--3842},
  year         = {2025}
}

@inproceedings{DBLP:conf/aaai/YalavarthiMSABJ24,
  author       = {Vijaya Krishna Yalavarthi and
                  Kiran Madhusudhanan and
                  Randolf Scholz and
                  Nourhan Ahmed and
                  Johannes Burchert and
                  Shayan Jawed and
                  Stefan Born and
                  Lars Schmidt{-}Thieme},
  title        = {{GraFITi}: Graphs for Forecasting Irregularly Sampled Time Series},
  booktitle    = {AAAI},
  pages        = {16255--16263},
  year         = {2024}
}

@inproceedings{DBLP:conf/icml/LiL0ZL025,
  author       = {Boyuan Li and
                  Yicheng Luo and
                  Zhen Liu and
                  Junhao Zheng and
                  Jianming Lv and
                  Qianli Ma},
  title        = {HyperIMTS: Hypergraph Neural Network for Irregular Multivariate Time
                  Series Forecasting},
  booktitle    = {ICML},
  year         = {2025}
}

@article{che2018recurrent,
  title={Recurrent neural networks for multivariate time series with missing values},
  author={Che, Zhengping and Purushotham, Sanjay and Cho, Kyunghyun and Sontag, David and Liu, Yan},
  journal={Scientific reports},
  volume={8},
  number={1},
  pages={6085},
  year={2018}
}

@inproceedings{DBLP:conf/iclr/ShuklaM21,
  author       = {Satya Narayan Shukla and
                  Benjamin M. Marlin},
  title        = {Multi-Time Attention Networks for Irregularly Sampled Time Series},
  booktitle    = {ICLR},
  year         = {2021}
}

@inproceedings{DBLP:conf/iclr/ZhangZTZ22,
  author       = {Xiang Zhang and
                  Marko Zeman and
                  Theodoros Tsiligkaridis and
                  Marinka Zitnik},
  title        = {Graph-Guided Network for Irregularly Sampled Multivariate Time Series},
  booktitle    = {ICLR},
  year         = {2022}
}

@inproceedings{DBLP:conf/kdd/ZhangZCBL23,
  author       = {Jiawen Zhang and
                  Shun Zheng and
                  Wei Cao and
                  Jiang Bian and
                  Jia Li},
  title        = {Warpformer: {A} Multi-scale Modeling Approach for Irregular Clinical
                  Time Series},
  booktitle    = {SIGKDD},
  pages        = {3273--3285},
  year         = {2023}
}

@inproceedings{DBLP:conf/nips/BrouwerSAM19,
  author       = {Edward De Brouwer and
                  Jaak Simm and
                  Adam Arany and
                  Yves Moreau},
  title        = {{GRU-ODE-Bayes}: Continuous Modeling of Sporadically-Observed Time Series},
  booktitle    = {NeurIPS},
  pages        = {7377--7388},
  year         = {2019}
}

@inproceedings{DBLP:conf/icml/HornMBRB20,
  author       = {Max Horn and
                  Michael Moor and
                  Christian Bock and
                  Bastian Rieck and
                  Karsten M. Borgwardt},
  title        = {Set Functions for Time Series},
  booktitle    = {ICML},
  pages        = {4353--4363},
  year         = {2020}
}

@inproceedings{DBLP:conf/nips/TashiroSSE21,
  author       = {Yusuke Tashiro and
                  Jiaming Song and
                  Yang Song and
                  Stefano Ermon},
  title        = {{CSDI:} Conditional Score-based Diffusion Models for Probabilistic
                  Time Series Imputation},
  booktitle    = {NeurIPS},
  pages        = {24804--24816},
  year         = {2021},
}

@inproceedings{mercatali2024graph,
  title={Graph neural flows for unveiling systemic interactions among irregularly sampled time series},
  author={Mercatali, Giangiacomo and Freitas, Andre and Chen, Jie},
  booktitle={NeurIPS},
  pages={57183--57206},
  year={2024}
}

@inproceedings{DBLP:conf/aaai/0003X0YZ00025,
  author       = {Chenxi Liu and
                  Qianxiong Xu and
                  Hao Miao and
                  Sun Yang and
                  Lingzheng Zhang and
                  Cheng Long and
                  Ziyue Li and
                  Rui Zhao},
  title        = {{TimeCMA}: Towards LLM-Empowered Multivariate Time Series Forecasting
                  via Cross-Modality Alignment},
  booktitle    = {AAAI},
  pages        = {18780--18788},
  year         = {2025}
}

@article{chang2026time,
  title={Time-imm: A dataset and benchmark for irregular multimodal multivariate time series},
  author={Chang, Ching and Hwang, Jeehyun and Shi, Yidan and Wang, Haixin and Wang, Wei and Peng, Wen-Chih and Chen, Tien-Fu},
  journal={Advances in Neural Information Processing Systems},
  volume={38},
  year={2026}
}

@inproceedings{DBLP:conf/iclr/0005WMCZSCLLPW24,
  author       = {Ming Jin and
                  Shiyu Wang and
                  Lintao Ma and
                  Zhixuan Chu and
                  James Y. Zhang and
                  Xiaoming Shi and
                  Pin{-}Yu Chen and
                  Yuxuan Liang and
                  Yuan{-}Fang Li and
                  Shirui Pan and
                  Qingsong Wen},
  title        = {Time-LLM: Time Series Forecasting by Reprogramming Large Language
                  Models},
  booktitle    = {ICLR},
  year         = {2024}
}

@inproceedings{DBLP:conf/icml/ZhongR0LW025,
  author       = {Siru Zhong and
                  Weilin Ruan and
                  Ming Jin and
                  Huan Li and
                  Qingsong Wen and
                  Yuxuan Liang},
  title        = {Time-VLM: Exploring Multimodal Vision-Language Models for Augmented
                  Time Series Forecasting},
  booktitle    = {ICML},
  year         = {2025}
}

@inproceedings{DBLP:conf/icde/LiuMXZLZLZ25,
  author       = {Chenxi Liu and
                  Hao Miao and
                  Qianxiong Xu and
                  Shaowen Zhou and
                  Cheng Long and
                  Yan Zhao and
                  Ziyue Li and
                  Rui Zhao},
  title        = {Efficient Multivariate Time Series Forecasting via Calibrated Language
                  Models with Privileged Knowledge Distillation},
  booktitle    = {ICDE},
  pages        = {3165--3178},
  year         = {2025}
}

@inproceedings{DBLP:conf/ijcai/0003ZX000025,
  author       = {Chenxi Liu and
                  Shaowen Zhou and
                  Qianxiong Xu and
                  Hao Miao and
                  Cheng Long and
                  Ziyue Li and
                  Rui Zhao},
  title        = {Towards Cross-Modality Modeling for Time Series Analytics: {A} Survey
                  in the {LLM} Era},
  booktitle    = {IJCAI},
  pages        = {10564--10572},
  year         = {2025}}

@inproceedings{silva2012predicting,
  title={Predicting in-hospital mortality of icu patients: The physionet/computing in cardiology challenge 2012},
  author={Silva, Ikaro and Moody, George and Scott, Daniel J and Celi, Leo A and Mark, Roger G},
  booktitle={Cinc},
  pages={245--248},
  year={2012}
}

@article{johnson2016mimic,
  title={MIMIC-III, a freely accessible critical care database},
  author={Johnson, Alistair EW and Pollard, Tom J and Shen, Lu and Lehman, Li-wei H and Feng, Mengling and Ghassemi, Mohammad and Moody, Benjamin and Szolovits, Peter and Anthony Celi, Leo and Mark, Roger G},
  journal={Scientific data},
  volume={3},
  number={1},
  pages={1--9},
  year={2016}
}

@article{vidulin2010localization,
  title={Localization data for person activity},
  author={Vidulin, Vedrana and Lustrek, Mitja and Kaluza, Bostjan and Piltaver, Rok and Krivec, Jana},
  journal={UCI Machine Learning Repository},
  volume={10},
  pages={C57G8X},
  year={2010}
}

@misc{menne2015long,
  title={Long-term daily climate records from stations across the contiguous United States},
  author={Menne, MJ and Williams Jr, CN and Vose, RS and Files, Data},
  year={2015}
}

@inproceedings{DBLP:conf/aaai/ZengCZ023,
  author       = {Ailing Zeng and
                  Muxi Chen and
                  Lei Zhang and
                  Qiang Xu},
  title        = {Are Transformers Effective for Time Series Forecasting?},
  booktitle    = {AAAI},
  pages        = {11121--11128},
  year         = {2023}
}

@inproceedings{DBLP:conf/iclr/WuHLZ0L23,
  author       = {Haixu Wu and
                  Tengge Hu and
                  Yong Liu and
                  Hang Zhou and
                  Jianmin Wang and
                  Mingsheng Long},
  title        = {TimesNet: Temporal 2D-Variation Modeling for General Time Series Analysis},
  booktitle    = {ICLR},
  year         = {2023}
}

@inproceedings{DBLP:conf/iclr/NieNSK23,
  author       = {Yuqi Nie and
                  Nam H. Nguyen and
                  Phanwadee Sinthong and
                  Jayant Kalagnanam},
  title        = {A Time Series is Worth 64 Words: Long-term Forecasting with Transformers},
  booktitle    = {ICLR},
  year         = {2023}
}

@inproceedings{DBLP:conf/iclr/ZhangY23,
  author       = {Yunhao Zhang and
                  Junchi Yan},
  title        = {Crossformer: Transformer Utilizing Cross-Dimension Dependency for
                  Multivariate Time Series Forecasting},
  booktitle    = {ICLR},
  year         = {2023}
}

@inproceedings{DBLP:conf/ijcai/WuPLJZ19,
  author       = {Zonghan Wu and
                  Shirui Pan and
                  Guodong Long and
                  Jing Jiang and
                  Chengqi Zhang},
  title        = {Graph WaveNet for Deep Spatial-Temporal Graph Modeling},
  booktitle    = {IJCAI},
  pages        = {1907--1913},
  year         = {2019}
}

@inproceedings{DBLP:conf/kdd/WuPL0CZ20,
  author       = {Zonghan Wu and
                  Shirui Pan and
                  Guodong Long and
                  Jing Jiang and
                  Xiaojun Chang and
                  Chengqi Zhang},
  title        = {Connecting the Dots: Multivariate Time Series Forecasting with Graph
                  Neural Networks},
  booktitle    = {SIGKDD},
  pages        = {753--763},
  year         = {2020}
}

@inproceedings{DBLP:conf/nips/CaoWDZZHTXBTZ20,
  author       = {Defu Cao and
                  Yujing Wang and
                  Juanyong Duan and
                  Ce Zhang and
                  Xia Zhu and
                  Congrui Huang and
                  Yunhai Tong and
                  Bixiong Xu and
                  Jing Bai and
                  Jie Tong and
                  Qi Zhang},
  title        = {Spectral Temporal Graph Neural Network for Multivariate Time-series
                  Forecasting},
  booktitle    = {NeurIPS},
  year         = {2020}
}

@inproceedings{DBLP:conf/nips/HuangSZDWZW23,
  author       = {Qihe Huang and
                  Lei Shen and
                  Ruixin Zhang and
                  Shouhong Ding and
                  Binwu Wang and
                  Zhengyang Zhou and
                  Yang Wang},
  title        = {CrossGNN: Confronting Noisy Multivariate Time Series Via Cross Interaction
                  Refinement},
  booktitle    = {NeurIPS},
  year         = {2023}
}

@article{DBLP:journals/corr/abs-2412-10302,
  author       = {Zhiyu Wu and
                  Xiaokang Chen and
                  Zizheng Pan and
                  Xingchao Liu and
                  Wen Liu and
                  others},
  title        = {DeepSeek-VL2: Mixture-of-Experts Vision-Language Models for Advanced
                  Multimodal Understanding},
  journal      = {CoRR},
  volume       = {abs/2412.10302},
  year         = {2024}
}

@article{DBLP:journals/corr/abs-2409-12191,
  author       = {Peng Wang and
                  Shuai Bai and
                  Sinan Tan and
                  Shijie Wang and
                  Zhihao Fan and
                  Jinze Bai and
                  Keqin Chen and
                  Xuejing Liu and
                  Jialin Wang and
                  Wenbin Ge and
                  Yang Fan and
                  Kai Dang and
                  Mengfei Du and
                  Xuancheng Ren and
                  Rui Men and
                  Dayiheng Liu and
                  Chang Zhou and
                  Jingren Zhou and
                  Junyang Lin},
  title        = {Qwen2-VL: Enhancing Vision-Language Model's Perception of the
                  World at Any Resolution},
  journal      = {CoRR},
  volume       = {abs/2409.12191},
  year         = {2024}
}

@inproceedings{DBLP:journals/corr/abs-2508-01971,
  author       = {Ziyu Zhou and
                  Yiming Huang and
                  Yanyun Wang and
                  Yuankai Wu and
                  James T. Kwok and
                  Yuxuan Liang},
  title        = {Revitalizing Canonical Pre-Alignment for Irregular Multivariate Time
                  Series Forecasting},
  booktitle     = {AAAI},
  year         = {2026}
}

@article{goldberger2000physiobank,
  title={PhysioBank, PhysioToolkit, and PhysioNet: components of a new research resource for complex physiologic signals},
  author={Goldberger, Ary L and Amaral, Luis AN and Glass, Leon and Hausdorff, Jeffrey M and Ivanov, Plamen Ch and Mark, Roger G and Mietus, Joseph E and Moody, George B and Peng, Chung-Kang and Stanley, H Eugene},
  journal={circulation},
  volume={101},
  number={23},
  pages={e215--e220},
  year={2000}
}

@inproceedings{DBLP:conf/aaai/Fan22,
  author       = {Jicong Fan},
  title        = {Dynamic Nonlinear Matrix Completion for Time-Varying Data Imputation},
  booktitle    = {AAAI},
  pages        = {6587--6596},
  year         = {2022}
}

@inproceedings{DBLP:conf/aaai/TangYSAMW20,
  author       = {Xianfeng Tang and
                  Huaxiu Yao and
                  Yiwei Sun and
                  Charu C. Aggarwal and
                  Prasenjit Mitra and
                  Suhang Wang},
  title        = {Joint Modeling of Local and Global Temporal Dynamics for Multivariate
                  Time Series Forecasting with Missing Values},
  booktitle    = {AAAI},
  pages        = {5956--5963},
  year         = {2020}
}

@article{scargle1982studies,
  title={Studies in astronomical time series analysis. II-Statistical aspects of spectral analysis of unevenly spaced data},
  author={Scargle, Jeffrey D},
  journal={Astrophys. J.},
  volume={263},
  pages={835--853},
  year={1982}
}

@inproceedings{DBLP:conf/icde/ZhangWYZXBW25,
  author       = {Yudong Zhang and
                  Xu Wang and
                  Xuan Yu and
                  Zhengyang Zhou and
                  Xing Xu and
                  Lei Bai and
                  Yang Wang},
  title        = {{DIFFODE:} Neural {ODE} with Differentiable Hidden State for Irregular
                  Time Series Analysis},
  booktitle    = {ICDE},
  pages        = {1--14},
  year         = {2025}
}

@inproceedings{DBLP:conf/iclr/OhLK24,
  author       = {YongKyung Oh and
                  Dongyoung Lim and
                  Sungil Kim},
  title        = {Stable Neural Stochastic Differential Equations in Analyzing Irregular
                  Time Series Data},
  booktitle    = {ICLR},
  year         = {2024}}

@inproceedings{DBLP:conf/icml/0008LSH23,
  author       = {Junnan Li and
                  Dongxu Li and
                  Silvio Savarese and
                  Steven C. H. Hoi},
  title        = {{BLIP-2:} Bootstrapping Language-Image Pre-training with Frozen Image
                  Encoders and Large Language Models},
  booktitle    = {ICML},
  volume       = {202},
  pages        = {19730--19742},
  year         = {2023},
}

@article{yi2023fouriergnn,
  title={FourierGNN: Rethinking multivariate time series forecasting from a pure graph perspective},
  author={Yi, Kun and Zhang, Qi and Fan, Wei and He, Hui and Hu, Liang and Wang, Pengyang and An, Ning and Cao, Longbing and Niu, Zhendong},
  journal={NeurIPS},
  volume={36},
  pages={69638--69660},
  year={2023}
}

@inproceedings{chenvisionts,
  title={VisionTS: Visual Masked Autoencoders Are Free-Lunch Zero-Shot Time Series Forecasters},
  author={Chen, Mouxiang and Shen, Lefei and Li, Zhuo and Wang, Xiaoyun Joy and Sun, Jianling and Liu, Chenghao},
  booktitle={ICML},
  year={2025}
}

@inproceedings{liu2026apn,
  title     = {Rethinking Irregular Time Series Forecasting: A Simple yet Effective Baseline},
  author    = {Liu, Xvyuan and Qiu, Xiangfei and Wu, Xingjian and Li, Zhengyu and Guo, Chenjuan and Hu, Jilin and Yang, Bin},
  booktitle = {AAAI},
  year      = {2026}
}
\endgroup

\appendix
\setcounter{figure}{0}
\setcounter{table}{0}
\renewcommand{\thefigure}{A\arabic{figure}}
\renewcommand{\thetable}{A\arabic{table}}
\section{Appendix}
\raggedbottom
\setlength{\textfloatsep}{6pt plus 1pt minus 1pt}
\setlength{\intextsep}{6pt plus 1pt minus 1pt}
\setlength{\floatsep}{6pt plus 1pt minus 1pt}
\subsection{Dataset Description}

Table~\ref{tab:dataset_statistics} summarizes the four main ISTS benchmark datasets used in the primary experiments. It reports the number of samples, variables, and the missing ratio. The image input scale quantities are reported separately in Section~\ref{sec:image_input_scale}.

\noindent\textbf{PhysioNet.}
The PhysioNet dataset~\citep{silva2012predicting} consists of 12,000 irregular multivariate time series collected from ICU patients within the first 48 hours after admission. We use the first 24h as observed data and the subsequent 24h for forecasting.

\noindent\textbf{MIMIC.}
MIMIC~\citep{johnson2016mimic} is a large-scale critical care database containing electronic health records of ICU patients. Following~\citep{DBLP:conf/nips/BilosSRJG21}, we extract 23,457 ISTS samples over 48-hour windows, using the first 24 hours for observation and the next 24 hours for prediction.

\noindent\textbf{Human Activity.}
The Human Activity dataset~\citep{vidulin2010localization} contains 12 irregular 3D positional variables from wearable sensors. We segment the original recordings into 5,400 ISTS samples, each spanning 4,000 milliseconds, where the first 3,000 milliseconds are used as observed data and the remaining 1,000 milliseconds are used for prediction.

\noindent\textbf{USHCN.}
USHCN~\citep{menne2015long} includes daily climate observations (5 variables) from U.S. meteorological stations. Following~\citep{DBLP:conf/nips/BrouwerSAM19}, we select 1996–2000 data from 1,114 stations. We chunk the dataset into 26,736 ISTS samples, using the first 24 months for observation to forecast conditions in the next month.

\subsection{Baseline Details}
\label{sec:baseline_details}
We compare our method with a diverse set of baselines spanning regular multivariate time series forecasting, irregularly sampled time series classification and imputation, and irregularly sampled time series forecasting.
\begin{itemize}[leftmargin=0.4cm]
    \item \textbf{DLinear}~\citep{DBLP:conf/aaai/ZengCZ023} decomposes time series into trend and remainder series, and models them with simple linear projections.
    \item \textbf{TimesNet}~\citep{DBLP:conf/iclr/WuHLZ0L23} captures multi-period temporal patterns by transforming one-dimensional sequences into two-dimensional representations.
    \item \textbf{PatchTST}~\citep{DBLP:conf/iclr/NieNSK23} adopts a Transformer architecture with patch-wise tokenization and channel independence to model long-term dependencies.
    \item \textbf{Crossformer}~\citep{DBLP:conf/iclr/ZhangY23} introduces cross-time and cross-dimension attention mechanisms to capture temporal and variable-wise interactions.
    \item \textbf{Graph WaveNet}~\citep{DBLP:conf/ijcai/WuPLJZ19} models inter-variable dependencies through a learnable adjacency matrix with diffusion convolution.
    \item \textbf{MTGNN}~\citep{DBLP:conf/kdd/WuPL0CZ20} jointly employs graph convolution and temporal convolution to learn dependencies across variables and time.
    \item \textbf{StemGNN}~\citep{DBLP:conf/nips/CaoWDZZHTXBTZ20} projects time series into the frequency domain via discrete Fourier transform and graph Fourier transform.
    \item \textbf{CrossGNN}~\citep{DBLP:conf/nips/HuangSZDWZW23} constructs multi-scale representations and applies cross-scale and cross-variable graph neural networks.
    \item \textbf{FourierGNN}~\citep{yi2023fouriergnn} builds a hypervariate graph and performs graph convolutions in the Fourier domain.
    \item \textbf{GRU-D}~\citep{che2018recurrent} extends GRU by incorporating time-decay mechanisms and missing-value handling strategies.
    \item \textbf{SeFT}~\citep{DBLP:conf/icml/HornMBRB20} represents time series as unordered sets and applies permutation-invariant set functions.
    \item \textbf{RainDrop}~\citep{DBLP:conf/iclr/ZhangZTZ22} leverages neural message passing and temporal self-attention to capture sensor dependencies.
    \item \textbf{Warpformer}~\citep{DBLP:conf/kdd/ZhangZCBL23} introduces a learnable warping module to align irregular time series at predefined temporal scales.
    \item \textbf{mTAND}~\citep{DBLP:conf/iclr/ShuklaM21} is an attention-based interpolation model that produces fixed-length representations from irregular sequences.
    \item \textbf{Latent-ODE}~\citep{DBLP:conf/nips/RubanovaCD19} defines continuous-time latent state dynamics using neural ordinary differential equations.
    \item \textbf{CRU}~\citep{DBLP:conf/icml/SchirmerELR22} combines Kalman filtering with an encoder-decoder framework for latent-state updates within ODE-based models.
    \item \textbf{Neural Flow}~\citep{DBLP:conf/nips/BilosSRJG21} parameterizes the solution trajectories of ordinary differential equations using neural networks.
    \item \textbf{T-PatchGNN}~\citep{DBLP:conf/icml/ZhangYL0024} transforms irregular time series into adaptive temporal patches and employs time-adaptive graph neural networks.
    \item \textbf{KAFNet}~\citep{DBLP:journals/corr/abs-2508-01971} combines sequence smoothing, learnable temporal compression, and frequency-domain linear attention.
    \item \textbf{APN}~\citep{liu2026apn} uses adaptive patching with time-aware patch aggregation to obtain regularized channel-independent representations from irregular observations.
    \item \textbf{Time-LLM}~\citep{DBLP:conf/iclr/0005WMCZSCLLPW24} uses textual prompts and reprogramming to adapt large language models to time series forecasting.
    \item \textbf{TimeCMA}~\citep{DBLP:conf/aaai/0003X0YZ00025} aligns time-series representations with textual prompt embeddings from large language models.
    \item \textbf{Time-VLM}~\citep{DBLP:conf/icml/ZhongR0LW025} explores vision-language models for time series forecasting by converting time series into visual and textual inputs.
    \item \textbf{ISTS-PLM}~\citep{DBLP:conf/kdd/0003Y0025} adapts pre-trained language models to ISTS with time-aware and variable-aware components.
\end{itemize}

\begin{table}[t]
\centering
\caption{Statistics of the main ISTS benchmark datasets.}
\label{tab:dataset_statistics}

\begin{tabularx}{\linewidth}{l*{3}{>{\centering\arraybackslash}X}}
\toprule
\textbf{Dataset} & \textbf{\#Samp.} & \textbf{\#Var.} & \textbf{Miss.\%} \\
\midrule
Human Activity & 5,400  & 12 & 75.0 \\
USHCN          & 26,736 & 5  & 77.9 \\
PhysioNet      & 12,000 & 41 & 88.4 \\
MIMIC-III      & 23,457 & 96 & 96.7 \\
\bottomrule
\end{tabularx}
\end{table}

\subsection{Implementation Details}
\label{sec:implementation_details}
All experiments are implemented in PyTorch and conducted on a Linux server equipped with an Intel(R) Xeon(R) Gold 6326 CPU (@ 2.90GHz) and an NVIDIA GeForce RTX 3090 GPU. We provide two MLLM backbone choices, DeepSeek-VL2-Tiny~\citep{DBLP:journals/corr/abs-2412-10302} and Qwen2-VL-2B-Instruct~\citep{DBLP:journals/corr/abs-2409-12191}; unless otherwise specified, Qwen2-VL-2B-Instruct is used as the default MLLM backbone for multimodal representation learning. To reduce computational overhead, all MLLM parameters are frozen during training, and only task-specific modules are optimized using Adam. Each experiment is repeated five times with different random seeds, and we report the mean and standard deviation. Following standard practice in ISTS forecasting, predictive performance is evaluated using Mean Squared Error (MSE) and Mean Absolute Error (MAE) on the test set.

\subsection{Additional Experimental Results}
\label{sec:supplementary_analysis}

This section provides extended analyses to further validate the robustness and efficiency of MM-ISTS.

\subsubsection{Image Input Scale}
\label{sec:image_input_scale}

For each sample, the Vision-Text Encoding branch constructs a three-channel image with spatial size $N\times L$, where $N$ is the number of variables and $L$ is the maximum historical input length. The image height corresponds to variables, and the width follows the historical input order. The three channels share this spatial layout and store values, masks, and time intervals. In Table~\ref{tab:image_scalability}, Pixels per Channel reports the original number of spatial positions in each constructed channel. Before the image is passed to the MLLM, we preprocess it according to the visual processor requirements of the corresponding MLLM. For DeepSeek-VL2-Tiny~\citep{DBLP:journals/corr/abs-2412-10302}, the constructed image is padded to $384\times384$ before model input. For Qwen2-VL-2B-Instruct~\citep{DBLP:journals/corr/abs-2409-12191}, each axis is padded to the nearest multiple of 28, following patch size 14 and merge size 2. The Qwen2-VL processor is initialized with a minimum pixels $28^2$ and a maximum of $1280\times28^2$ pixels. The listed datasets have $N\le96$ and $L\le292$, so their constructed images are smaller than $384\times384$ and remain far below the maximum pixel budget used for Qwen2-VL-2B-Instruct. Therefore, preprocessing does not crop variables or historical steps, so the information of each irregular time series sample is fully preserved.

\begin{table}[t]
\centering
\caption{Constructed image sizes for each dataset.}
\label{tab:image_scalability}
\setlength{\tabcolsep}{2.5pt}
\begin{tabularx}{\linewidth}{l|l *{3}{>{\centering\arraybackslash}X}}
\toprule
\textbf{Group} & \textbf{Dataset} & \textbf{Vars. ($N$)} & \textbf{Max Hist. ($L$)} & \textbf{Pixels/Ch. ($N\times L$)} \\
\midrule
\multirow{4}{*}{\shortstack[l]{Main\\benchmarks}} & Human Activity & 12 & 98  & 1,176 \\
& USHCN          & 5  & 205 & 1,025 \\
& PhysioNet      & 41 & 128 & 5,248 \\
& MIMIC-III      & 96 & 280 & 26,880 \\
\midrule
\multirow{6}{*}{\shortstack[l]{Additional\\datasets with\\thier own texts}} & FNSPID       & 6  & 23  & 138 \\
& ClusterTrace & 11 & 228 & 2,508 \\
& GDELT        & 5  & 292 & 1,460 \\
& CESNET       & 10 & 159 & 1,590 \\
& RepoHealth   & 10 & 31  & 310 \\
& EPA-Air      & 4  & 168 & 672 \\
\bottomrule
\end{tabularx}
\end{table}

\begin{figure}[!htbp]
  \centering
  \begin{subfigure}[b]{0.48\linewidth}
    \centering
    \includegraphics[width=\linewidth]{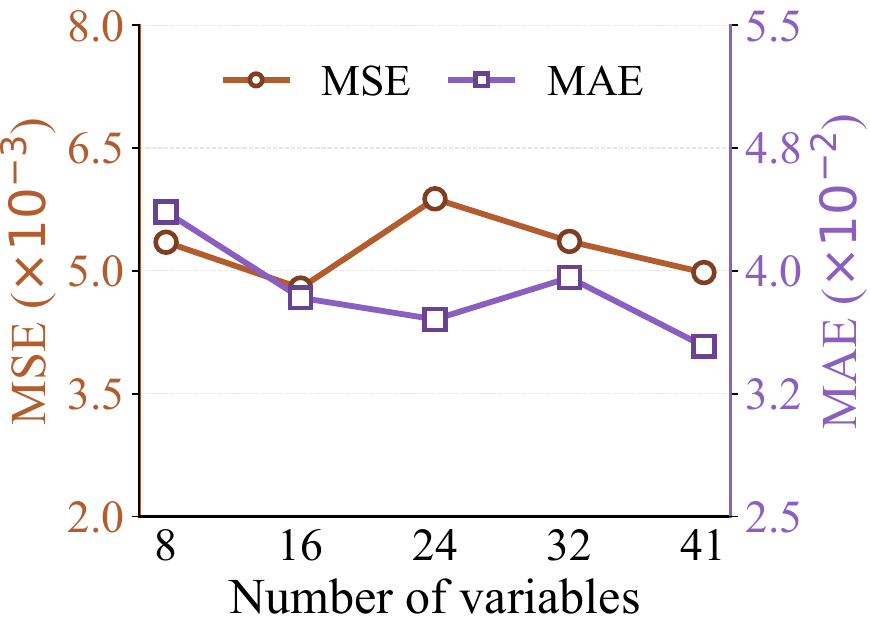}
    \caption{Performance}
  \end{subfigure}
  \hfill
  \begin{subfigure}[b]{0.48\linewidth}
    \centering
    \includegraphics[width=\linewidth]{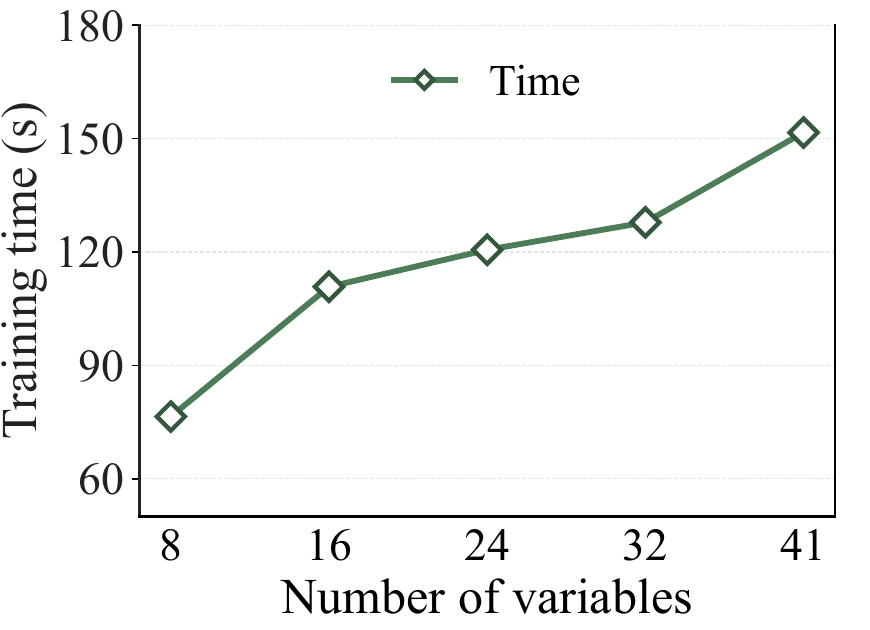}
    \caption{Training time}
  \end{subfigure}
  \caption{Effect of different numbers of variables on PhysioNet.}
  \label{fig:variable_scalability}

\end{figure}

We evaluate MM-ISTS on PhysioNet with different numbers of variables under the same forecasting setting. To further examine the effect of image input scale, we use a fixed random seed to sample 20\%, 40\%, 60\%, and 80\% of the variables, and use all variables in the full PhysioNet setting. Figure~\ref{fig:variable_scalability}~(a) shows that the prediction errors vary only slightly as the number of variables increases, suggesting stable forecasting performance. Figure~\ref{fig:variable_scalability}~(b) shows that the training time per epoch increases gradually with more variables. The growth is smooth and roughly linear, without a sharp rise in time cost. These results indicate that MM-ISTS maintains stable forecasting performance under different variable counts, while the added training cost grows moderately.

\subsubsection{Hidden Layer of MLLM}
\label{sec:hidden_layer_appendix}


We investigate the impact of extracting hidden states from different layers of the Qwen2-VL-2B-Instruct~\citep{DBLP:journals/corr/abs-2409-12191}. As shown in Table~\ref{tab:vlm_layer}, we evaluate four layer positions counting backward from the final layer on Human Activity~\citep{vidulin2010localization}. The results reveal that the 3rd-to-last layer achieves the best overall performance, with the lowest MSE and a tied best MAE. Interestingly, extracting features from the last layer does not yield optimal results. This phenomenon can be attributed to the fact that the last layer of MLLMs is typically optimized for next-token prediction in language generation.
In contrast, the intermediate layers retain richer and more generalizable multimodal semantic representations. Furthermore, we observe that extracting from earlier layers (e.g., 7th-to-last) leads to performance degradation, likely because cross-modal fusion in transformer-based MLLMs becomes more complete in deeper layers. Notably, despite these variations, all layer configurations yield comparable results with only minor fluctuations, indicating that MM-ISTS is robust to the choice of hidden layer and that the MLLM representations are also beneficial across different depths.

\begin{table}[t]
\centering
\caption{Effect of different MLLM hidden layers.}
\vspace{-0.3cm}
\label{tab:vlm_layer}
\setlength{\tabcolsep}{5.2mm}
\begin{tabular}{c|c|c}
\toprule
\textbf{Layer Position} & \textbf{MSE$\times 10^{-3}$} & \textbf{MAE$\times 10^{-2}$} \\
\midrule
Last              & $2.49 \pm 0.03$ & $3.08 \pm 0.03$ \\
3rd-to-last       & $\mathbf{2.47} \pm \mathbf{0.01}$ & $\mathbf{3.06} \pm \mathbf{0.02}$ \\
5th-to-last       & $2.48 \pm 0.02$ & $3.06 \pm 0.02$ \\
7th-to-last       & $2.49 \pm 0.02$ & $3.07 \pm 0.03$ \\
\bottomrule
\end{tabular}

\end{table}

\subsubsection{Hyperparameter Sensitivity Analysis}
\label{sec:hyperparameter_appendix}
\begin{figure}[t]
    \centering
    \begin{subfigure}[b]{0.49\columnwidth}
        \centering
        \includegraphics[width=\linewidth]{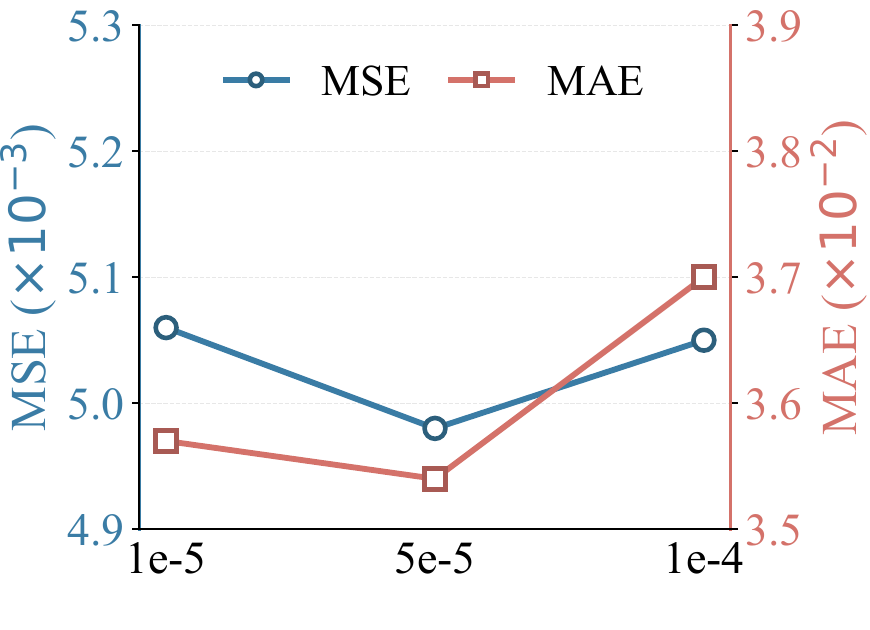}
        \caption{Learning rate}
    \end{subfigure}
    \hfill
    \begin{subfigure}[b]{0.49\columnwidth}
        \centering
        \includegraphics[width=\linewidth]{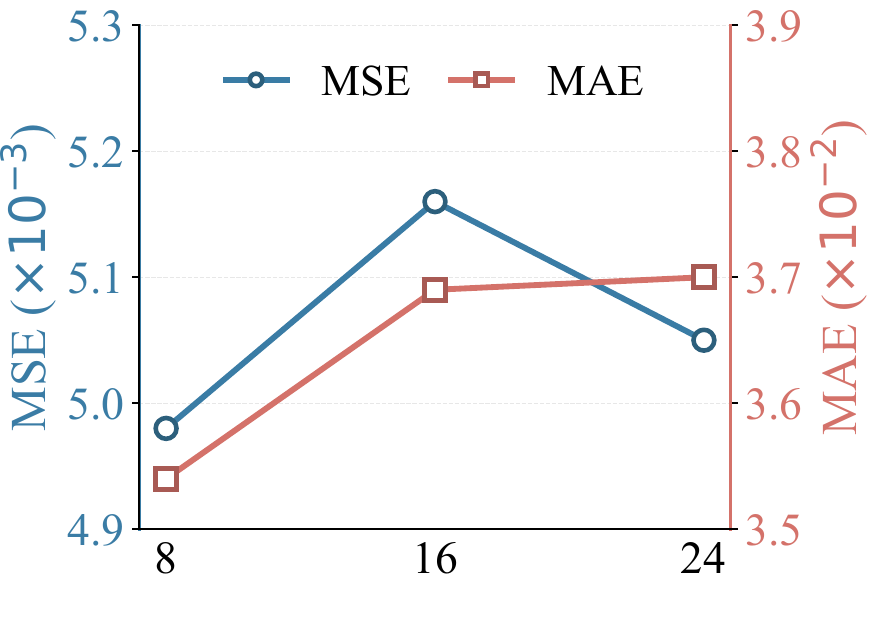}
        \caption{Batch size}
    \end{subfigure}

    \vspace{0.5mm}

    \begin{subfigure}[b]{0.49\columnwidth}
        \centering
        \includegraphics[width=\linewidth]{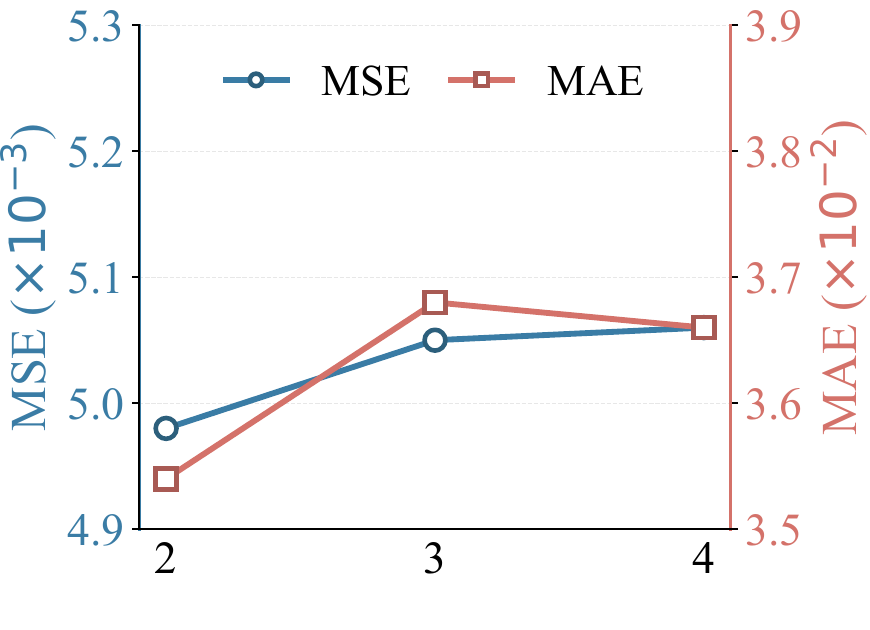}
        \caption{QBE layers}
    \end{subfigure}
    \hfill
    \begin{subfigure}[b]{0.49\columnwidth}
        \centering
        \includegraphics[width=\linewidth]{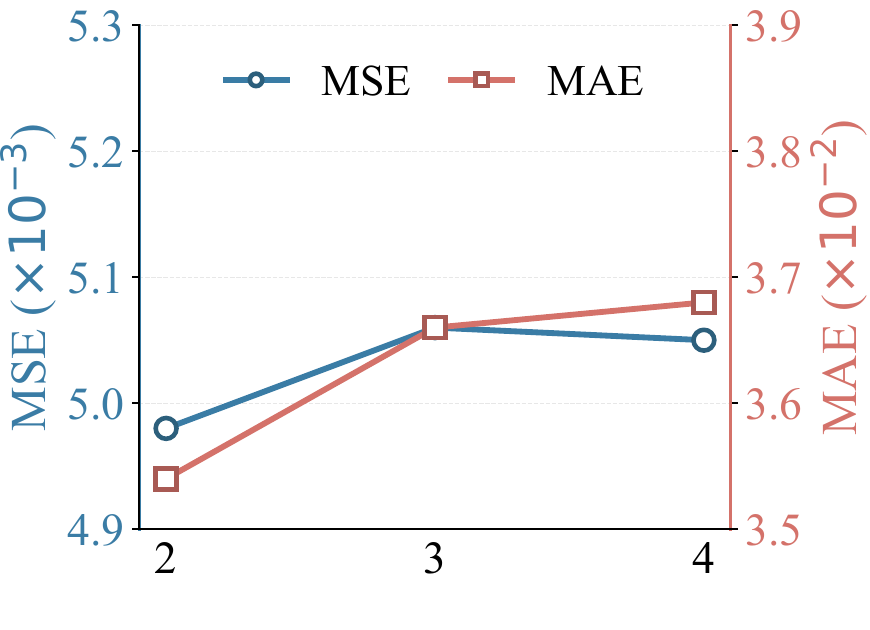}
        \caption{TE/VE layers}
    \end{subfigure}

    \vspace{0.5mm}

    \begin{subfigure}[b]{0.49\columnwidth}
        \centering
        \includegraphics[width=\linewidth]{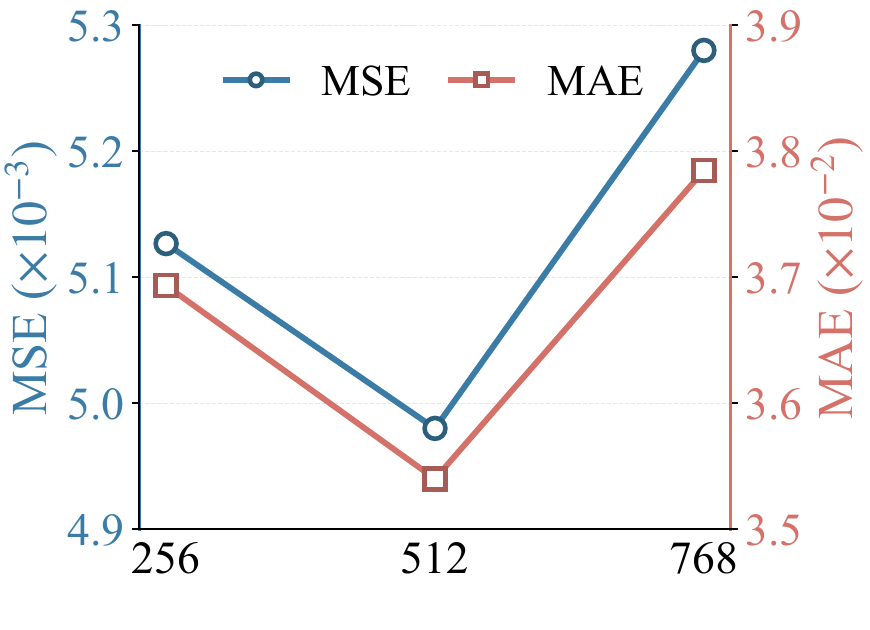}
        \caption{Hidden dimension}
    \end{subfigure}
    \hfill
    \begin{subfigure}[b]{0.49\columnwidth}
        \centering
        \includegraphics[width=\linewidth]{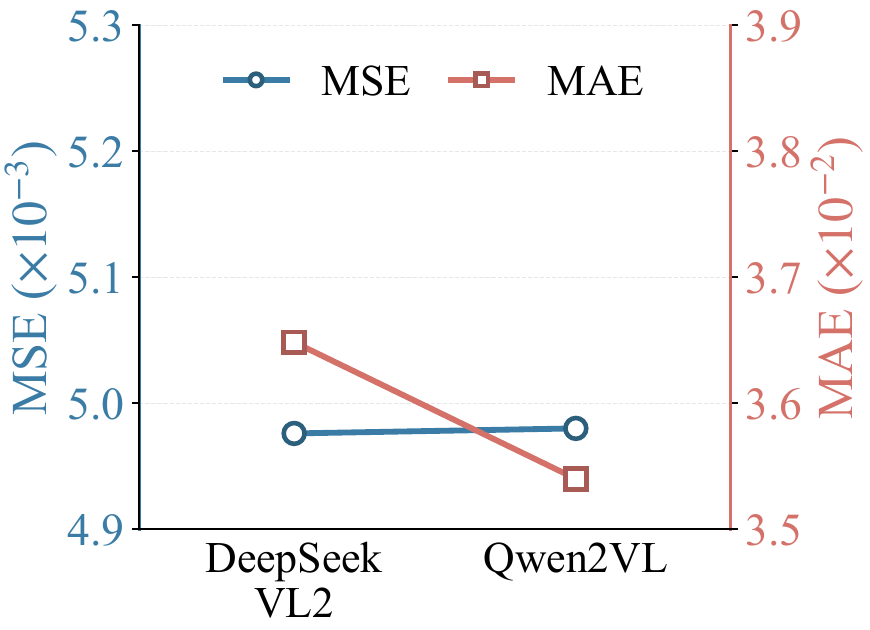}
        \caption{MLLM backbone}
    \end{subfigure}
    \caption{PhysioNet hyperparameter sensitivity.}
    \label{fig:all_hyperparams}
\end{figure}

\begin{figure}[t]
    \centering
    \begin{subfigure}[b]{0.49\columnwidth}
        \centering
        \includegraphics[width=\linewidth]{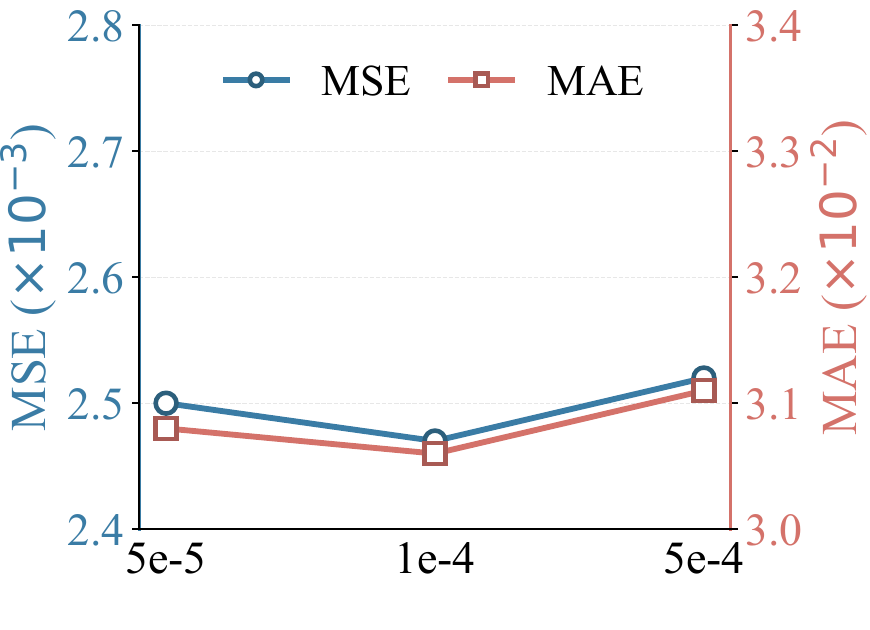}
        \caption{Learning rate}
    \end{subfigure}
    \hfill
    \begin{subfigure}[b]{0.49\columnwidth}
        \centering
        \includegraphics[width=\linewidth]{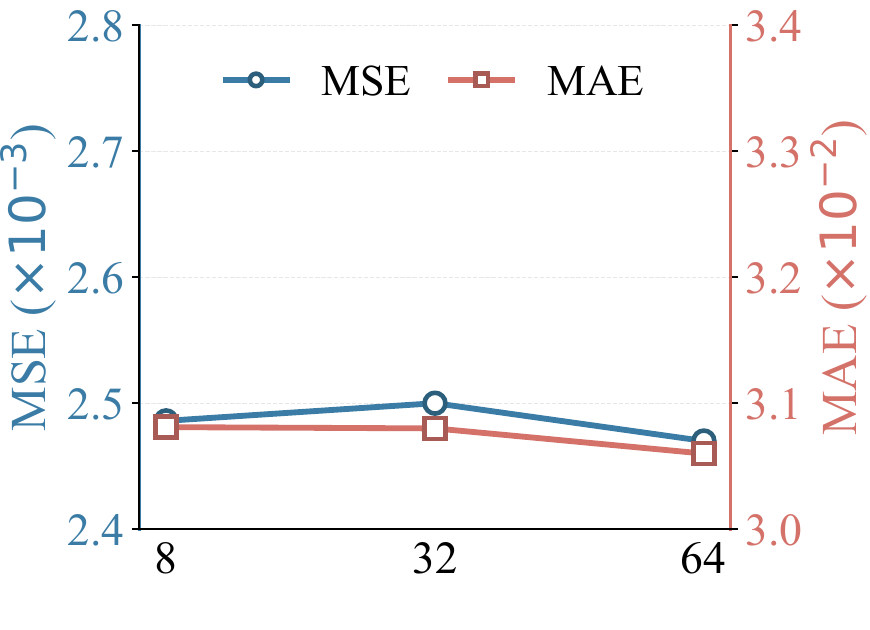}
        \caption{Batch size}
    \end{subfigure}

    \vspace{0.5mm}

    \begin{subfigure}[b]{0.49\columnwidth}
        \centering
        \includegraphics[width=\linewidth]{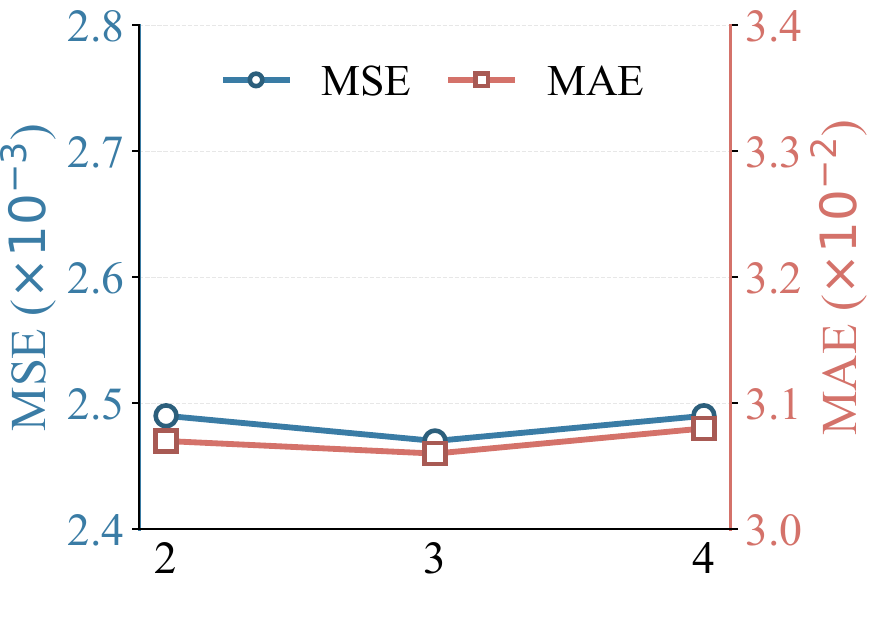}
        \caption{QBE layers}
    \end{subfigure}
    \hfill
    \begin{subfigure}[b]{0.49\columnwidth}
        \centering
        \includegraphics[width=\linewidth]{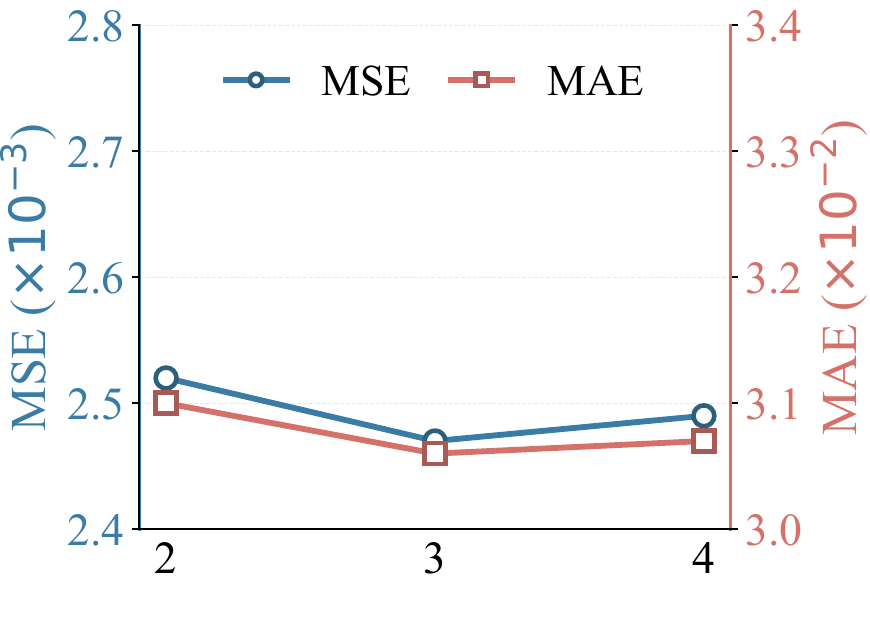}
        \caption{TE/VE layers}
    \end{subfigure}

    \vspace{0.5mm}

    \begin{subfigure}[b]{0.49\columnwidth}
        \centering
        \includegraphics[width=\linewidth]{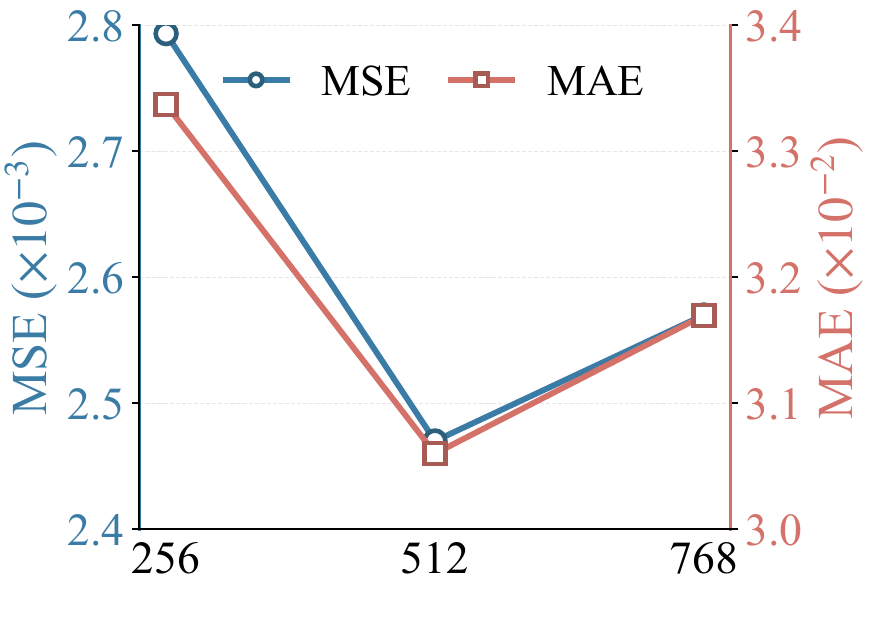}
        \caption{Hidden dimension}
    \end{subfigure}
    \hfill
    \begin{subfigure}[b]{0.49\columnwidth}
        \centering
        \includegraphics[width=\linewidth]{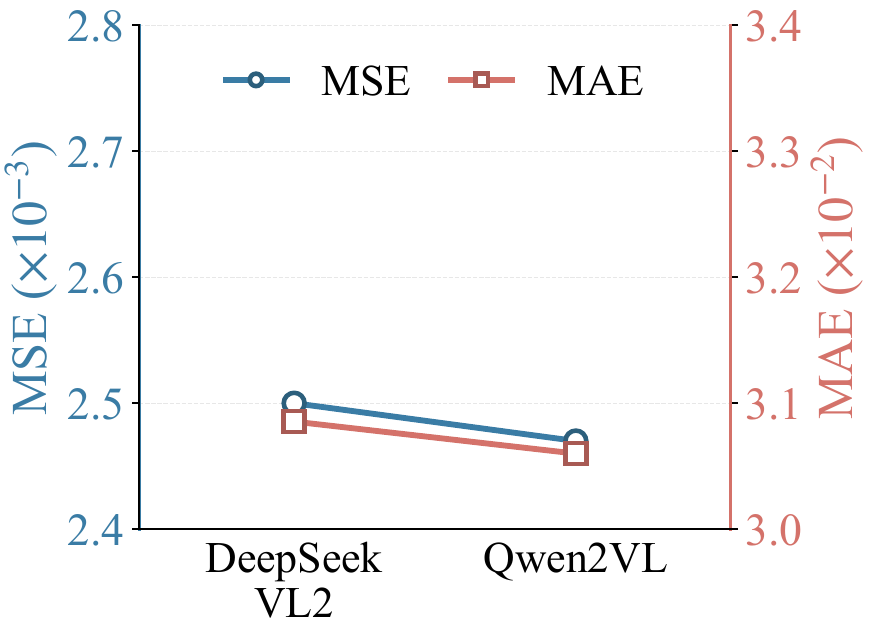}
        \caption{MLLM backbone}
    \end{subfigure}
    \caption{Human Activity hyperparameter sensitivity.}
    \label{fig:hyperparams_activity}
\end{figure}

Figure~\ref{fig:all_hyperparams} investigates the sensitivity of MM-ISTS to key hyperparameters on the PhysioNet dataset. We vary the learning rate, batch size, the number of layers in the Adaptive Query-Based Feature Extractor, the depths of Temporal Encoder and Variable Encoder, the hidden feature dimension, and the MLLM backbone. Among the displayed settings, $5\times10^{-5}$ gives the lowest error for learning rate, batch size 8 gives the lowest error for batch size, the two-layer setting gives the lowest errors for both encoder depths, $D=512$ gives the best hidden-dimension result, and the two MLLM backbones show comparable forecasting performance. Figure~\ref{fig:hyperparams_activity} reports the sensitivity and configuration analysis on Human Activity. For learning rate, we compare $\{5\times10^{-5}, 10^{-4}, 5\times10^{-4}\}$, and $10^{-4}$ performs best. For the Adaptive Query-Based Feature Extractor layers and TE/VE layers, we test $\{2, 3, 4\}$. The three-layer setting gives the best result for both parts of Human Activity. We also include hidden feature dimension and MLLM backbone selection in this analysis. $D=512$ remains the best hidden-dimension setting, and Qwen2-VL-2B-Instruct~\citep{DBLP:journals/corr/abs-2409-12191} gives slightly lower errors than DeepSeek-VL2-Tiny~\citep{DBLP:journals/corr/abs-2412-10302} on this dataset. Figure~\ref{fig:human_activity_efficiency_heatmap}~(a) further reports the interaction between learning rate and batch size on Human Activity. Overall, nearby settings remain close, indicating that MM-ISTS is not overly sensitive to small changes around the selected configuration.

\subsubsection{Additional Efficiency Analysis}
\label{sec:physionet_efficiency_appendix}
\begin{figure}[!htbp]
  \centering
  \includegraphics[width=0.55\linewidth]{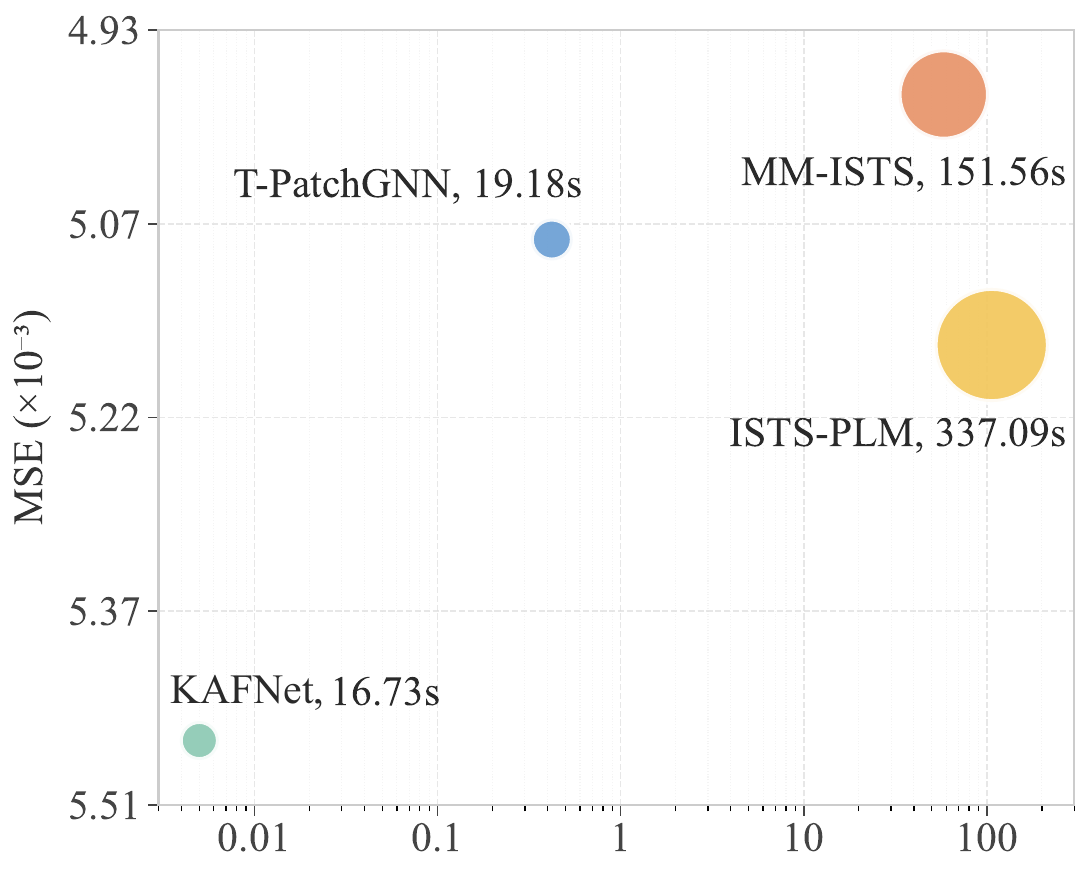}
  \caption{Efficiency analysis on PhysioNet.}
  \label{fig:physionet_efficiency_appendix}
\end{figure}

Figure~\ref{fig:physionet_efficiency_appendix} provides an additional efficiency comparison on PhysioNet, complementing the Human Activity result in Figure~\ref{fig:human_activity_efficiency_heatmap}~(b). The same overall trend can be observed: MM-ISTS requires more training time per epoch than compact ISTS models such as KAFNet and T-PatchGNN, but it achieves lower forecasting error. Compared with ISTS-PLM, the most relevant LLM-based baseline, MM-ISTS, uses substantially less training time per epoch while maintaining stronger prediction performance. This result further supports the efficiency benefit of freezing the MLLM backbone and using compact query embeddings for multimodal alignment.

\subsubsection{ISTS Encoding Branch Component Analysis}
\begin{figure}[!htbp]
  \centering
  \begin{subfigure}[b]{0.48\linewidth}
    \centering
    \includegraphics[width=\linewidth]{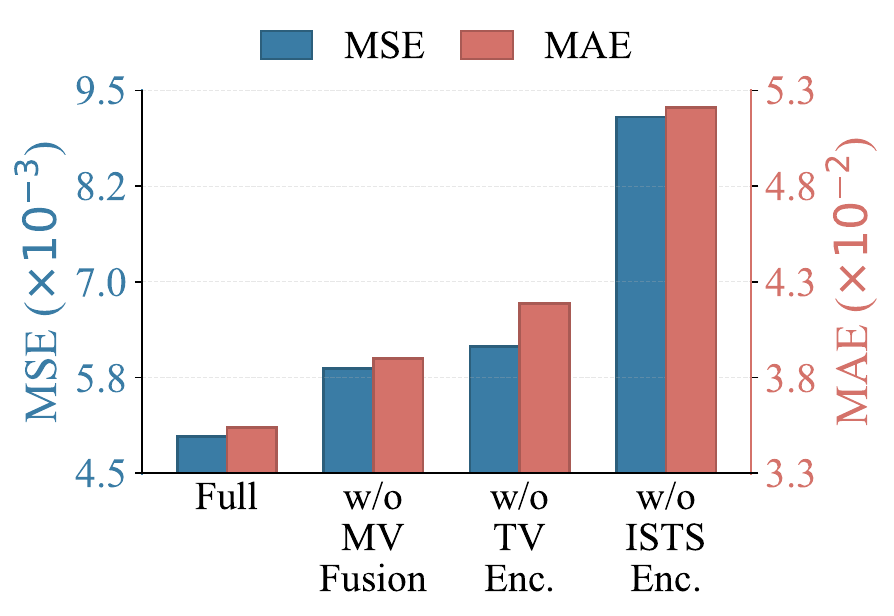}
    \caption{PhysioNet}
  \end{subfigure}
  \hfill
  \begin{subfigure}[b]{0.48\linewidth}
    \centering
    \includegraphics[width=\linewidth]{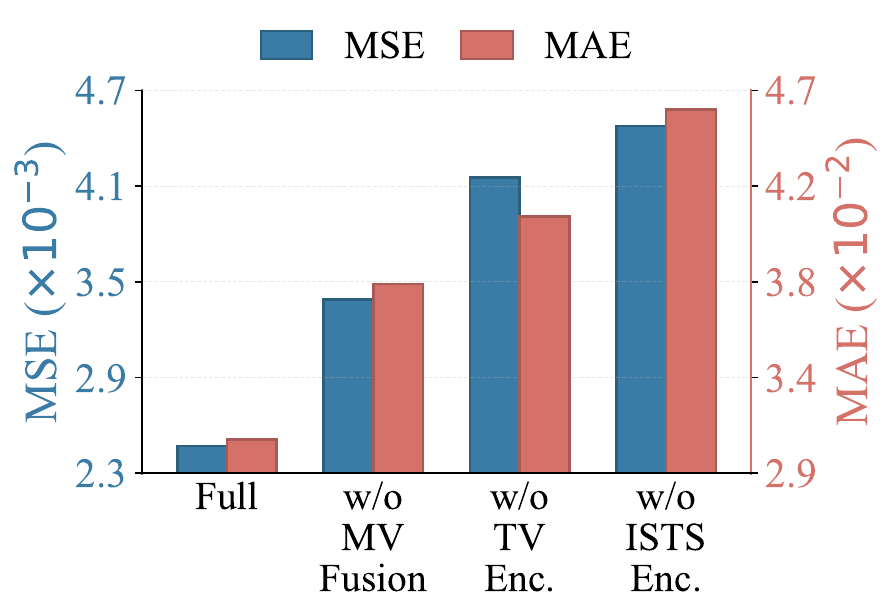}
    \caption{Human Activity}
  \end{subfigure}
  \caption{ISTS Encoding branch component ablation study.}
  \label{fig:ists_branch_ablation}
\end{figure}

We further examine the contribution of the ISTS Encoding branch on PhysioNet and Human Activity. The full model results are the same as those reported in Table~\ref{tab:main_results} and Figure~\ref{fig:ablation1}. Figure~\ref{fig:ists_branch_ablation} compares the full model with three variants. \textit{w/o MV Fusion} removes Multi-View Embedding Fusion, which combines timestamp, value, mask, and variable embeddings before temporal modeling. \textit{w/o TV Enc.} removes the Temporal-Variable Encoder, which models temporal patterns within each variable and dependencies across variables. \textit{w/o ISTS Enc.} removes the whole ISTS Encoding branch and keeps only the multimodal branch for prediction. Removing the whole ISTS Encoding branch causes the largest degradation on both datasets. This indicates that the ISTS Encoding branch remains necessary even when MLLM features are available. Removing Multi-View Embedding Fusion also increases the error, showing that the ISTS Encoding branch benefits from combining different ISTS embeddings before temporal modeling. Removing the Temporal-Variable Encoder gives a larger drop than removing the fusion module on Human Activity and also hurts PhysioNet, suggesting that temporal modeling within variables and dependency modeling across variables are both important for accurate ISTS forecasting under irregular sampling.

\subsubsection{T-SNE Analysis}
\begin{figure}[!t]
    \centering
    \includegraphics[width=\linewidth]{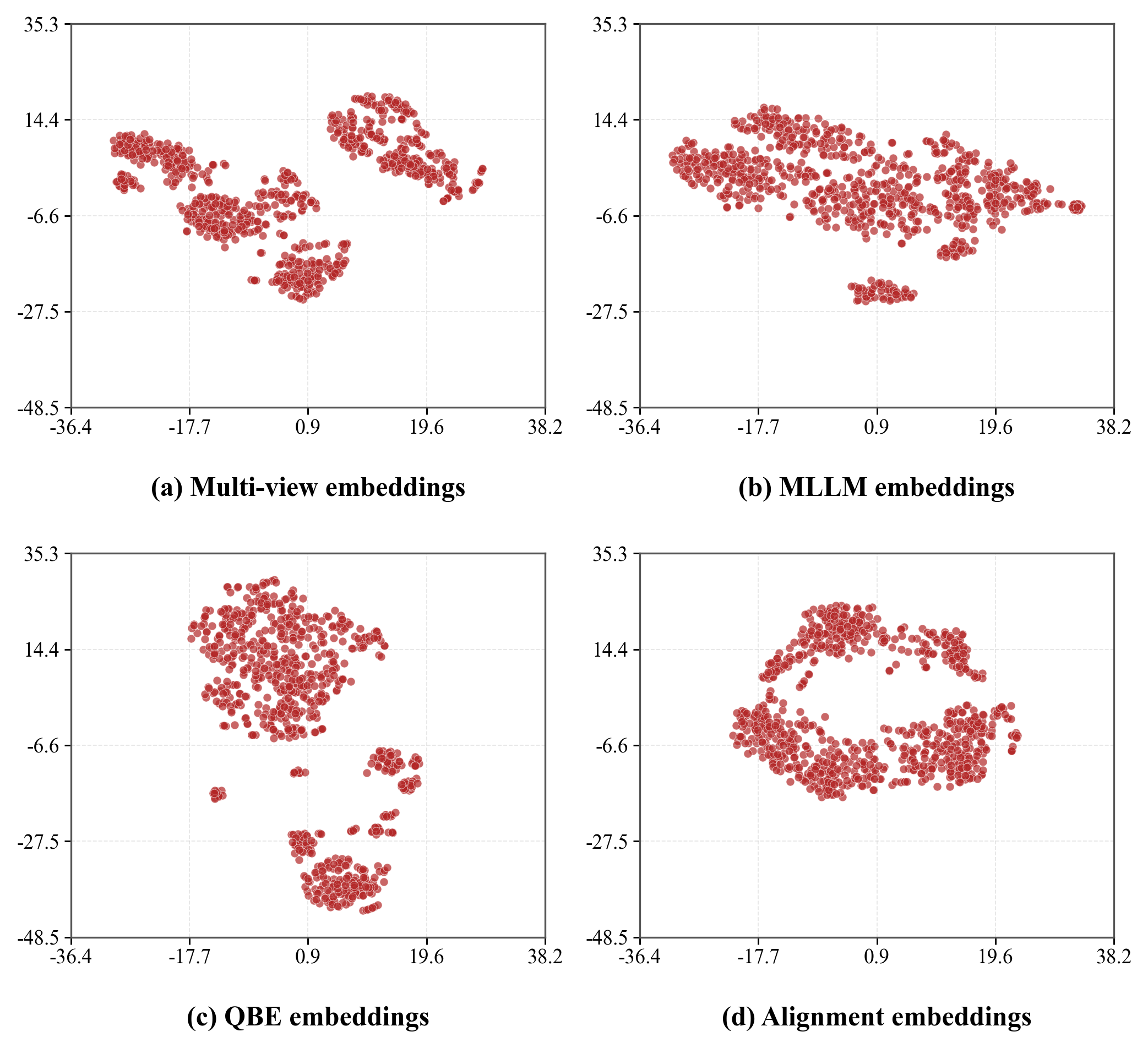}
    \caption{T-SNE visualization of representations from four model stages.}
    \label{fig:tsne_visualization}
\end{figure}

Figure~\ref{fig:tsne_visualization} visualizes four embeddings collected during inference: fused embeddings from Multi-View Embedding Fusion, mean-pooled MLLM token embeddings, Adaptive Query-Based Feature Extractor embeddings, and multimodal alignment embeddings. The early fused embeddings and raw MLLM token embeddings show relatively large overlap. After the Adaptive Query-Based Feature Extractor and multimodal alignment, the points form clearer groups and local structures. This visualization suggests that the later modules make the learned representations more organized for the final forecasting task.

\subsection{Algorithm}
\label{sec:algorithm}

Algorithm~\ref{alg:mm_ists} summarizes the MM-ISTS pipeline. Lines 2--4 construct the irregularity-aware image and text prompt, and then encode them with the frozen MLLM. Lines 6--9 embed the observed ISTS values, timestamps, and masks, and produce numerical variable-level representations. Line 11 applies the Adaptive Query-Based Feature Extractor to convert MLLM hidden states into variable-aligned multimodal embeddings. Lines 13--17 compute observation statistics, estimate modality-aware gating weights, apply cross-attention between numerical and multimodal features, and generate the final forecasts from the fused representation.

\begin{algorithm}[H]
\caption{MM-ISTS Pipeline}
\label{alg:mm_ists}
\begingroup
\setlength{\parindent}{0pt}
\begin{flushleft}
\textbf{Input:} Observed values $\mathbf{X}_{\mathit{obs}} \in \mathbb{R}^{B \times L \times N}$, timestamps $\mathbf{T}_{\mathit{obs}} \in \mathbb{R}^{B \times L \times N}$, observation mask $\mathbf{M}_{\mathit{obs}} \in \{0,1\}^{B \times L \times N}$, prediction times $\mathbf{T}_{\mathit{pred}} \in \mathbb{R}^{B \times L_{\mathit{pred}}}$, dataset description $\mathcal{D}$, and learned queries $\mathbf{Q} \in \mathbb{R}^{N \times d_m}$.\\[-0.1em]
\textbf{Output:} Forecasts $\hat{\mathbf{X}}_{\mathit{pred}} \in \mathbb{R}^{B \times L_{\mathit{pred}} \times N}$.
\end{flushleft}
\begin{algorithmic}[1]
\STATE \textcolor{gray}{\textit{// Phase 1: Cross-Modal Vision-Text Encoding}}
\STATE $\mathcal{I} \gets \mathit{Visualize}(\mathbf{X}_{\mathit{obs}}, \mathbf{T}_{\mathit{obs}}, \mathbf{M}_{\mathit{obs}})$;
\STATE $\mathcal{P} \gets \mathit{GeneratePrompt}(\mathbf{X}_{\mathit{obs}}, \mathcal{D}, \mathbf{M}_{\mathit{obs}})$;
\STATE $\mathbf{E}_{\mathit{MLLM}} \gets \mathit{MLLM}(\mathcal{I}, \mathcal{P})$;

\STATE \textcolor{gray}{\textit{// Phase 2: ISTS Encoding}}
\STATE $\mathbf{E}, \mathbf{E}_{\mathit{var}} \gets \mathit{Embed}(\mathbf{X}_{\mathit{obs}}, \mathbf{T}_{\mathit{obs}}, \mathbf{M}_{\mathit{obs}})$;
\STATE $\mathbf{H}_{\mathit{time}} \gets \mathit{TemporalEncoder}(\mathbf{E})$;
\STATE $\mathbf{H}_{\mathit{time}} \gets \mathit{MaskPool}(\mathbf{H}_{\mathit{time}}, \mathbf{M}_{\mathit{obs}})$;
\STATE $\mathbf{H}_{\mathit{ISTS}} \gets \mathit{VariableEncoder}(\mathbf{H}_{\mathit{time}} + \mathbf{E}_{\mathit{var}})$;

\STATE \textcolor{gray}{\textit{// Phase 3: Adaptive Query-Based Feature Extraction}}
\STATE $\mathbf{H}_{\mathit{MM}} \gets \mathit{QBE}(\mathbf{Q}, \mathbf{E}_{\mathit{MLLM}})$;

\STATE \textcolor{gray}{\textit{// Phase 4: Multimodal Alignment}}
\STATE $\mathbf{s} \gets [\mu(\mathbf{X}_{\mathit{obs}}), \sigma(\mathbf{X}_{\mathit{obs}}), \rho, c]$;
\STATE $\alpha^{\mathit{num}}, \alpha^{\mathit{mm}} \gets \mathit{Softmax}(\mathit{GatingNet}(\mathbf{s}))$;
\STATE $\mathbf{H}_{\mathit{fused}} \gets \mathit{CrossAttention}(\mathbf{H}_{\mathit{ISTS}}, \mathbf{H}_{\mathit{MM}}, \mathbf{H}_{\mathit{MM}})$;
\STATE $\mathbf{H}_{\mathit{final}} \gets \alpha^{\mathit{num}}\mathbf{H}_{\mathit{ISTS}} + \alpha^{\mathit{mm}}\mathbf{H}_{\mathit{fused}}$;
\STATE $\hat{\mathbf{X}}_{\mathit{pred}} \gets \mathit{Predictor}(\mathbf{H}_{\mathit{final}}, \mathbf{T}_{\mathit{pred}})$;

\RETURN $\hat{\mathbf{X}}_{\mathit{pred}}$;
\end{algorithmic}
\endgroup
\end{algorithm}

\end{document}